\documentclass[10pt,twocolumn,letterpaper]{article}

\usepackage{wacv}              

\usepackage{graphicx}
\usepackage{amsmath}
\usepackage{amssymb}
\usepackage{booktabs}

\usepackage{url}

\usepackage{makecell}
\usepackage{caption}
\usepackage{array}
\usepackage{multirow, wrapfig}
\usepackage{tabularx}

\usepackage{algorithm}
\usepackage{algpseudocode}

\usepackage{arydshln}

\usepackage[pagebackref,breaklinks,colorlinks]{hyperref}
\usepackage[accsupp]{axessibility} 

\usepackage[capitalize]{cleveref}
\crefname{section}{Sec.}{Secs.}
\Crefname{section}{Section}{Sections}
\Crefname{table}{Table}{Tables}
\crefname{table}{Tab.}{Tabs.}

\def\wacvPaperID{144} 
\def\confName{WACV}
\def\confYear{2025}

\algnewcommand{\LineComment}[1]{\State \(\triangleright\) #1}

\newtoggle{papermain}
\toggletrue{papermain}

\begin{document}

\title{Zero-Shot Class Unlearning in CLIP with Synthetic Samples} 

\author{Alexey Kravets\\
University of Bath, \\
Bath, UK\\
{\tt\small ak3095@bath.ac.uk}
\and
Vinay P. Namboodiri\\
University of Bath, \\
Bath, UK \\
{\tt\small vpn22@bath.ac.uk}
}
\maketitle


\begin{abstract}
Machine unlearning is a crucial area of research. It is driven by the need to remove sensitive information from models to safeguard individuals' right to be forgotten under rigorous regulations such as GDPR. In this work, we focus on unlearning within CLIP, a dual vision-language encoder model trained on a massive dataset of image-text pairs using contrastive loss. To achieve forgetting we expand the application of Lipschitz regularization to the multimodal context of CLIP. Specifically, we smooth both visual and textual embeddings associated with the class intended to be forgotten relative to the perturbation introduced to the samples from that class. Additionally, importantly, we remove the necessity for real forgetting data by generating synthetic samples via gradient ascent maximizing the target class. Our forgetting procedure is iterative, where we track accuracy on a synthetic forget set and stop when accuracy falls below a chosen threshold. We employ a selective layers update strategy based on their average absolute gradient value to mitigate over-forgetting. We validate our approach on several standard datasets and provide thorough ablation analysis and comparisons with previous work. 
\end{abstract}


\section{Introduction}
\label{sec:intro}

\begin{figure}[!t]
\centering
\includegraphics[width=1.\linewidth]{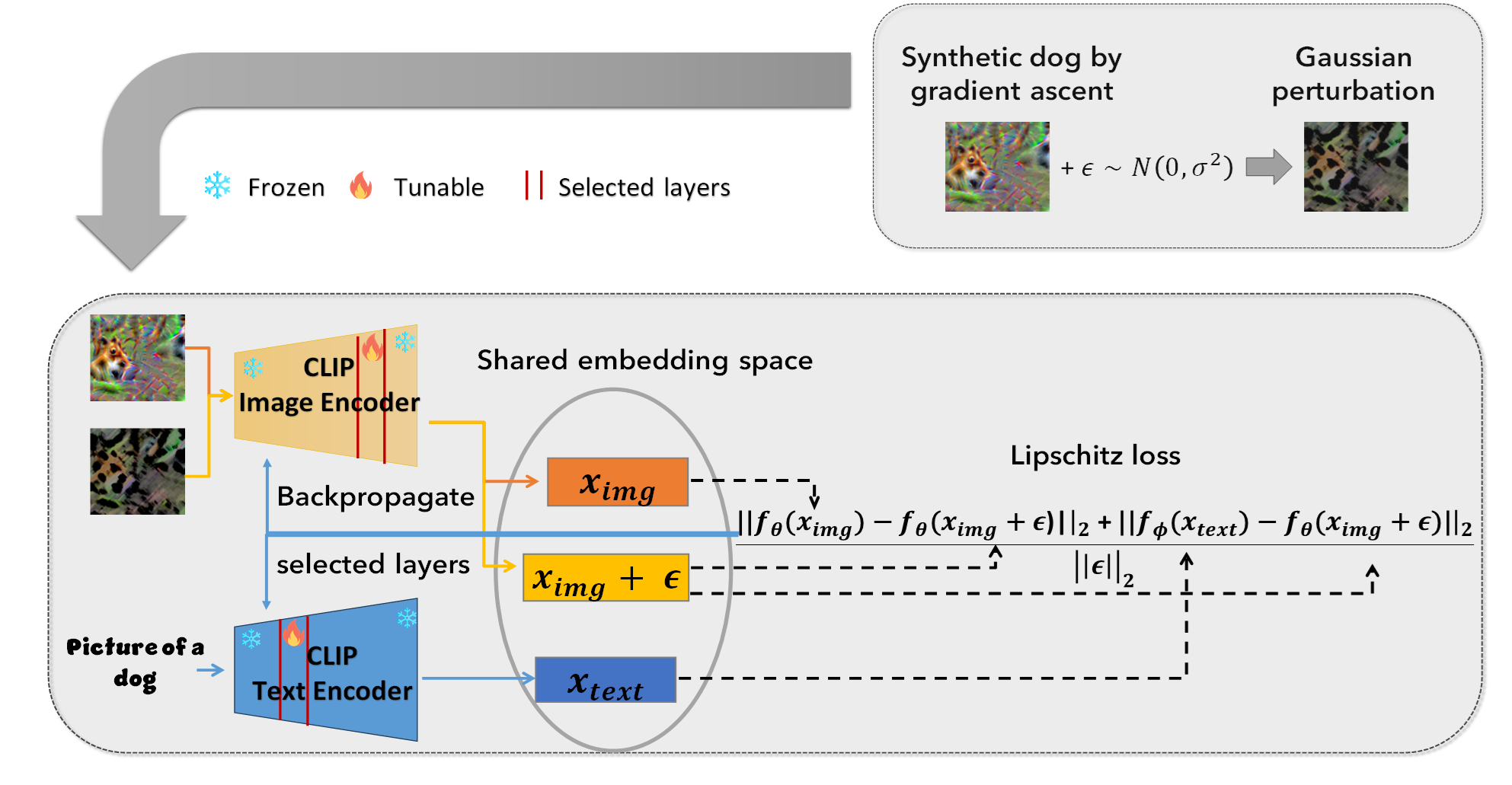}
\caption{\textbf{Overview of the approach.} First, we generate synthetic images of a class to forget by gradient ascent. Then, we perform a Gaussian perturbation of the images and pass the original and perturbed images through CLIP image encoder and textual description of the class through CLIP textual encoder. As image and text are projected into a shared embedding space, we can use the final representation of the perturbed image as a perturbed representation of text "Picture of a dog". Finally, we apply Lipschitz regularization and backpropagate to selected layers based on their importance to visual and textual encoders.}
\label{fig:fig_method}
\end{figure}

\textbf{Motivation} Machine unlearning \cite{unlearning_survey} is becoming an important research area as the need to remove specific learned information from models grows. CLIP~\cite{radford_2021_learning}, a versatile model utilized in fields like robotics, content moderation, and image classification, plays a critical role in many systems. However, this widespread use also raises concerns. If CLIP has inadvertently learned to recognize sensitive or proprietary information during its training, this knowledge can spread across different applications leading to serious ethical and legal issues. For instance, if private images accidentally leaked into the model during training, causing it to recognize individuals, regulations such as the GDPR\footnote{https://gdpr-info.eu/art-17-gdpr/} mandate that such information must be erased. Additionally, in dynamic applications like object tracking or image classification, where models are regularly retrained or updated with new data, the ability to selectively unlearn outdated or incorrect information without needing to retrain the entire model from scratch could save considerable computational resources and ensure that the model remains accurate and up-to-date.

\textbf{Challenges of Unlearning CLIP:} CLIP and other related vision-language models can be used for image and text-based tasks without being retrained. The unlearning for such models is especially challenging for the following three reasons: a) they have separate visual and textual encoders. Thus, even if we unlearn using the image representation, the textual modality may still be able to generalise. For instance, if we want to forget the concept of a specific breed of dog - Chihuahua and are able to use a method to unlearn in the image representation, the model may still be able to caption the dog Chihuahua as the information may be retained in the textual representation b) we do not have access to the data used to train the CLIP model\footnote{https://github.com/openai/CLIP/issues/127}. Techniques that require availability of samples for unlearning are thus not applicable c) CLIP model has a large number of parameters hence retraining it is unfeasible. In our work, we explicitly design a method that overcomes all these challenges.

\textbf{Devising a solution:} In order to solve the problem, we need a basic principle that we can use for unlearning. The most straightforward method is to retrain the original model without using specific data that is intended to be forgotten. As mentioned, for a foundational model like CLIP, that is computationally expensive. The other approach could be to use random sampling of data to be retained and some samples of the data to be forgotten and using a technique like amnesiac forgetting \cite{graves_2020_amnesiac} to adapt the CLIP model. As CLIP model spans many concepts, that is not practical. We next consider methods that rely only on samples of data to be forgotten. A recent method \cite{foster_2024_zeroshot} shows that by retraining the model by perturbing the samples to be forgotten, one can achieve unlearning. We follow this approach and aim to extend it to address all the challenges associated with unlearning in CLIP.

The first step in using the perturbation approach is to determine the type of perturbation to apply. We compared Lipschitz with other embedding perturbation methods and found that it performs well. Consequently, we adopted this approach for unlearning. Next, we tackled the challenge of needing real data samples for forgetting. Since we lacked access to the actual data used in training the model, we addressed this problem by generating synthetic image samples using gradient ascent~\cite{szegedy_2013_intriguing}. We further solved the problem of unlearning with the dual modalities noticing that a simple extension of \cite{foster_2024_zeroshot} adding continuous Gaussian noise to discrete textual tokens to unlearn the textual encoder is not possible; thus we propose to use the image representation as a proxy for text representation to unlearn the textual encoder. We finally needed to solve the computational challenge of unlearning. We do so by selectively changing the weights of only a few layers instead of retraining the whole model. We pursued an iterative unlearning approach that allows us to precisely control the amount of unlearning required to forget the specific class while retaining the information for all other classes.

\textbf{Overview of our approach:} We provide an overview of our approach in Fig. \ref{fig:fig_method}. Let us consider that a specific class needs to be forgotten, such as the class of dog \textit{Chihuahua}. Our approach involves generating synthetic samples of this class using gradient ascent. We then use iterative forgetting through Lipschitz regularization using the synthetic image samples and do the same for the textual representation through the joint image-text representation. \\

\textbf{Contributions}
Our main contributions are summarizes as follows: a) differently from previous methods \cite{foster_2024_zeroshot} our approach does not rely on any real training data to forget, thus it is truly zero-shot. b) We extend Lipschitz unlearning to CLIP in a novel way noticing that adding continuous Gaussian noise to discrete textual tokens to unlearn the textual encoder is not possible. c) We propose an iterative forgetting procedure with a clear stopping criteria based on the accuracy of the synthetic samples used for forgetting.

\section{Related Work}

\paragraph{Multimodal Unlearning}
Multimodal forgetting remains underexplored in the literature. Authors in \cite{cheng_2023_multimodal} introduce multimodal unlearning defining it by three key properties: modality decoupling, unimodal knowledge retention, and multimodal knowledge retention. The methodology involves optimizing a multimodal model through three losses to effectively unlearn forgotten data while preserving the knowledge of retained data, satisfying these properties. However, this methodology cannot be applied to CLIP due to its non-parametric fusion of modalities. Also the method requires training data for knowledge retention. \\
In \cite{zhang_2023_forgetmenot}, authors attempt to induce forgetting in Stable Diffusion (SD) through attention steering. This process entails minimizing cross-attention maps from the Stable Diffusion model between latent input features and textual embeddings of concepts intended for forgetting. This disentangles textual associations from image associations of target concepts. Similarly for the SD model, authors in \cite{gandikota_2023_erasing} utilize the inverse of the energy-based composition \cite{ho_2022_classifierfree} to guide generation probability away from conditional towards the unconditional prediction of concepts to be forgotten by negating the predicted noise associated with the forget concept. 
The techniques used for generative models are not applicable to dual-encoder models such as CLIP. To the best of our knowledge, none of the existing multimodal unlearning methods specifically address forgetting in CLIP.

\paragraph{Machine Unlearning with Generated Data}

In \cite{tarun_2022_fast}, the authors generated anti-samples for classes meant to be forgotten by error maximization, which is the reverse of the minimization process employed during training. These anti-samples exhibit patterns opposite to those of the sample classes. They also train using data samples from the training data classes to be remembered. 
During the CLIP forgetting procedure, we also generate synthetic data. However, our approach diverges from that in \cite{tarun_2022_fast} as our method employs loss minimization to generate synthetic data with same patterns rather than opposite patterns of the forget data. 
The challenge induced by the approach \cite{tarun_2022_fast} is that a delicate balance needs to be struck between retraining with classes to be remembered vs anti-samples of classes that need to be forgotten. Our approach uses regularization and synthetic samples in a more efficient manner.

\paragraph{Zero-shot Machine Unlearning}

Authors in \cite{chundawat_2023_zeroshot} have taken the approach in \cite{tarun_2022_fast} a step further by eliminating the dependence on real data from the classes meant to be retained. They utilize a synthesis approach similar to that in \cite{tarun_2022_fast} for generating forget data, while the retain data are synthesized using an error minimization procedure. This solution is, therefore, zero-shot as it does not rely on any real data. Note that this method is not as practical as our approach, as for general models like CLIP, there is no explicit class that needs to be retained, but, the general capabilities needs to be retained. 
The work that is closest to ours is the work by Foster {\it et. al.} \cite{foster_2024_zeroshot}. They perform forgetting via local Lipschitz regularization on unimodal vision models. They do not require the retain data but rely on real data to be forgotten for training which does not make them completely zero-shot. We extended their method to multimodal CLIP model and eliminate the need for actual data. We provide comparisons with both \cite{foster_2024_zeroshot} and \cite{chundawat_2023_zeroshot} and show that our proposed method performs better for CLIP while being more practical than either of these methods.  

\paragraph{Non-zero-shot Machine Unlearning Methods}
SalUn\cite{fan2023salun} unlearns models by updating parameters based on weight saliency, needing real retain and forget data. Unrolling SGD \cite{sgdunl} reverses gradient updates to forget specific data points, requiring an SGD-trained model, unlike CLIP’s Adam training. \cite{WarPirWreRie20} introduces a method for selectively removing features and labels from models based on the concept of influence functions but also requires real data.

\section{Preliminaries}
\label{section:background}

\subsection{CLIP Dual Encoder Model}
CLIP is a dual encoder model that consists of visual and textual components. The visual component processes images and extracts their features while the textual component processes textual descriptions and encodes them into a fixed-length vector representation. Due to the contrastive pre-training during which CLIP learns to associate images and text in a shared embedding space, CLIP is able to perform a variety of zero-shot tasks, including classification. 

For classification, given \textit{N} classes, they are encoded within a contextual prompt such as \textit{"A photo of a \{class\}"} with the CLIP textual encoder $f_\phi$. This results in the classifier weight matrix $W \in \mathbb{R}^{N\times d}$, where \textit{d} represents the embedding dimension. When presented with a test image $I_i$, it is encoded using the CLIP image encoder $f_\theta$:
\begin{equation}
    T_i = f_\theta(I_i), \ T_i \in \mathbb{R}^{d}.
\end{equation}
Following this encoding step, the dot product between the matrix $W$ and the embedded image $T_i$ is computed to get the zero-shot classification logits for the image $I_i$:
\begin{equation}
    \text{CLIPlogits}_i = T_i W^T, \ \text{CLIPlogits}_i \in  \mathbb{R}^{N}.
\label{eq:clip_logits}
\end{equation}

\subsection{Local Lipschitz Regularization}
In order to unlearn samples for a particular class, we rely on an existing technique. Our method is based on work by Foster {\it et. al.}\cite{foster_2024_zeroshot} that utilizes the concept of Lipschitz continuity for forgetting. The idea is to locally perturb an input image that needs to be forgotten by a Gaussian noise and minimizing the ratio between the change in the outputs for the perturbed and unperturbed images and change in the input. This regularization was first proposed by Yoshida and Miyato \cite{yoshida_2017_spectral} as a means for obtaining generalization. However, \cite{foster_2024_zeroshot} observed that using sufficient Gaussian perturbation, the learnt response for the particular input is unlearnt. Formally, given some Gaussian noise $\boldsymbol{\epsilon} \sim \mathcal{N}(0,\sigma^2)$ of the same dimensionality of the input image $\boldsymbol{x}$ we minimize:
\begin{equation}
    \ell=\mathbb{E}\left(\frac{\left\|f_\theta(\boldsymbol{x})-f_\theta(\boldsymbol{x}+\boldsymbol{\epsilon})\right\|_2}{\|\epsilon\|_2}\right). 
\end{equation} 
The expectation is approximated by averaging over $N$ perturbations. We also evaluated direct regularization on the embedding loss and observed the Lipschitz regularization to be better. We discuss it in more details in sec. \ref{sec:results}.

Lipschitz regularization while being a useful way for unlearning, can be a weak signal not removing enough class information from the model. Our approach therefore relies on assuring that we ensure the unlearning of the class by validating on a set of examples synthesized for the class and using iterative unlearning for good performance.

\section{Method}
\label{section:approach}

\subsection{Setting}
Given the model's training set denoted as $D$, in a standard machine unlearning setting we identify two subsets of $D$: data to retain $D_r$ and data to forget $D_f$. However, our setting differs because the training data for CLIP have not been made publicly available. Consequently, we are unable to determine whether a specific data sample was used for training. Therefore, we cannot verify the forgetting of a particular sample or compare the forgetting performance to a retrained model on $D_r$ excluding $D_f$. 
Even if we had access to the training data, achieving the latter would be infeasible due to the substantial computational resources required to retrain large-scale models like CLIP. 

\subsection{Extending Lipschitz Regularization to CLIP}
Authors in \cite{foster_2024_zeroshot} utilize local Lipschitz regularization exclusively on vision models. Our experiments reveal that updating solely the vision branch is insufficient for CLIP to forget a selected class and adjustments to both vision and text branches are necessary. Adapting Lipschitz regularization to a dual encoder CLIP model poses a challenge as there is no direct method to perturb discrete language tokens with Gaussian noise. One potential approach involves directly modifying tokens, however, determining the degree of noise introduced to the input becomes ambiguous. Since the final layer embedding from both the CLIP vision and language branches are mapped to a shared image-text space, we can avoid perturbing the text directly. Instead, we use the perturbed image as a proxy for the perturbed text and compute Lipschitz regularization for the text branch in the same manner as for the image branch. Thus, we define our loss objective for both vision and text branches as follows:

\begin{align}
    \ell &= \mathbb{E}\left(\frac{\left\|f_\theta(\boldsymbol{x_{img}}) - f_\theta(\boldsymbol{x_{img}} + \boldsymbol{\epsilon})\right\|_2}{\|\epsilon\|_2} \right. \\
    &\quad \left. + \frac{\|f_\phi(\boldsymbol{x_{text}}) - f_\theta(\boldsymbol{x_{img}} + \boldsymbol{\epsilon})\|_2}{ \|\epsilon\|_2} \right),
\label{eq:main_loss}
\end{align}

where $f_\theta$ is the image encoder and $f_\phi$ is the text encoder that output last layers embeddings, $x_{img}$ is the image sample and $x_{text}$ its corresponding class wrapped in the contextual prompt. As the CLIP objective ensures that the shared embeddings for image and text are close, requiring the text embedding to be close to the perturbed image embedding is a valid unlearning regularization. The expectation in the equation \ref{eq:main_loss} is approximated with Monte Carlo using $N$ perturbations for each sample.   

\subsection{Synthetic Forget Samples}
We create synthetic forget samples by performing gradient ascent to maximize the target class \cite{szegedy_2013_intriguing}. Starting from random noise sample we perform the following update until the prediction on $x$ is of a desired class:
\begin{equation}
    x = x + \alpha \frac{\partial L(x, y)}{\partial x},
\label{eq:synt_samples}
\end{equation}

where $\alpha$ is the learning rate, $y$ the desired target class and $L$ the loss function. We then use these synthetic forget samples to update the weights in CLIP using local Lipschitz regularization loss. 
These synthetic samples do not have the appearance of the sample class (examples in Appendix \ref{app:synth_vis}) due to the simple approach we use for generation, however, these samples suffice for unlearning a class. Additional details about data generation are found in Appendix \ref{app:forg_success}.

\subsection{Layers Update Based on the Average Gradient}

We observed that updating all parameters of CLIP results in excessive forgetting. Therefore, we perform a selective update of layers based on their importance to the samples we aim to forget. To determine this importance, we calculate the average absolute gradient value of the layers and update a specific number of layers in both the vision and text branches during each iteration. Results of the ablation on forgetting with all parameters are shown in Tab. \ref{table:aggregated_ablations}. 

\subsection{Stopping Criteria}

To achieve gradual forgetting, we begin with a low value of $\sigma$ and a small number of layers to update in CLIP. During forgetting we monitor the accuracy on the synthetic samples and stop the forgetting process when it falls below a predefined threshold. If during forgetting the accuracy of the synthetic data does not drop we increase both the amount of noise $\sigma$ and the number of layers for a more aggressive forgetting. Full algorithm is shown in the Appendix \ref{app:forg_algo}.

\section{Experiments}
\label{section:experiments}

\subsection{Comparable Methods}
As our approach is the first to be proposed for unlearning in CLIP, there are no direct comparable methods available. Therefore, we have adapted a number of methods to provide a fair evaluation of our approach. These are:

\paragraph{L2 embedding regularization loss (Emb)}
\label{par:embedreg}
Similarly to the method outlined before we perturb the inputs with a Gaussian noise but this time, instead of Lipschitz regularization we utilize L2 regularization that is defined as follows:
\begin{align}
M & = \mathbb{E}(\left\|f_\theta(\boldsymbol{x_{img}})-f_\theta(\boldsymbol{x_{img}}+\boldsymbol{\epsilon})\right\|_2 + \\ & 
\|f_\phi(\boldsymbol{x_{text}})-
f_\theta(\boldsymbol{x_{img}}+\boldsymbol{\epsilon})\|_2) + \; 
\\ & 
\alpha\cdot \|f_\theta(\boldsymbol{x_{img}}) + f_\phi(\boldsymbol{x_{text})}\|_2 .
\end{align}
As only synthetic forget data is used it is a zero-shot method like ours (denoted \textit{ZS} in the results Tab. \ref{table:forget_synth_aggreg}). 
\paragraph{Amnesiac forgetting with synthetic data (Amns)}
We adapt the approach from \cite{graves_2020_amnesiac} to the multimodal setting fine-tuning CLIP with the contrastive loss. We replace the labels corresponding to the forget class randomly with a different label using synthetic data. To keep it zero-shot we do not use the retain data but only train with the forget data. 
\paragraph{Error Minimization-Maximization Noise (EMMN)}
We adapt the method in \cite{chundawat_2023_zeroshot} to multimodal setting learning retain and forget samples through loss minimization and maximization respectively and train the model on them. As the method does not require any real data it is zero-shot. 
\paragraph{Unimodal Lipschitz (ULip)}
We perform forgetting only on the visual encoder of CLIP as in \cite{foster_2024_zeroshot} using image perturbation and local Lipschitz regularization. We run the method using \textbf{real} data to forget. As it requires real data it is not completely zero-shot (denoted \textit{semi ZS} in Tab. \ref{table:forget_synth_aggreg}).
\paragraph{Amnesiac forgetting with real data including retain data (AmnsRetain)}
Similarly to \textit{Amns} \cite{graves_2020_amnesiac} described earlier we replace the labels corresponding to the forget class randomly with a different label using \textbf{real} data. This time we include the retain real data from the dataset to which the label to forget belongs to. As this method uses the data to retain it is not zero-shot (denoted \textit{not ZS} in Tab. \ref{table:forget_synth_aggreg}).
\paragraph{SalUn} SalUn \cite{fan2023salun} utilizes the forget data to compute the weights saliency which are used to select the parameters to update enabling unlearning. We extend SalUn to CLIP using the version of SalUn for image classification with random labeling as in the paper. This method is also \textit{not ZS}.

\subsection{Datasets}

We assess CLIP's forgetting using four high-quality fine-grained datasets:
Caltech101 \cite{feifei_2007_learning} consists of images belonging to 101 distinct categories containing examples of objects or scenes. StanfordCars \cite{krause_2013_3d} includes images of cars categorized into different makes and models. OxfordFlowers \cite{nilsback_2008_automated} comprises images of flowers from 102 species while StanfordDogs \cite{khosla_novel} contains images of dogs categorized into different breeds. These datasets comprise images spanning various categories with minimal overlap between them, thus we do not need to filter for similar classes to the forget class across different datasets during evaluation.

\subsection{Implementation Details}
We perform our experiments on CLIP with ResNet50 \cite{he_2015_deep} as visual encoder. Experiments with ViT \cite{dosovitskiy_2020_an} and other implementation details can be found in Appendix \ref{app:vit_results} and \ref{app:add_impl} respectively. 

\subsection{Evaluation}

As we mentioned previously, we are unable to compare CLIP performance on a forget class to the retrained version of the model without the forget data as we effectively do not know whether a certain sample was used to train CLIP due to the data being not open sourced. However, even if the data were open sourced the computational power required to retrain such a big model that relies on a huge amount of data for its zero-shot capabilities would be a challenge. Therefore, after the forgetting procedure we will evaluate CLIP's classification performance on the selected class for forgetting, the remaining classes from that dataset and the classification performance on the remaining three datasets. \\
It's important to highlight that we aim for the accuracy on the forget class to be as low as possible, while maintaining similar accuracy levels on the remaining classes of dataset the forget class belongs to and all other datasets compared to before the application of the forgetting procedure. We summarize this information in one number for an easier comparison computing the average score. To calculate it we compute the normalized reduction in accuracy for the target class to be forgotten, denoted as $A_{cl}$, and the normalized reduction in accuracy for the other classes, denoted as $A_{ds}$, across the $N$ datasets examined. Average score is then computed as follows:
\begin{equation}
    \text{Avg. Score} = \frac{1}{N+1} ((1 - A_{cl}) + \sum_{ds} A_{ds}).
\end{equation}
This score varies between 0 in case unlearning is complete with accuracy of the target class of 0 while maintaining all not targeted classes accuracy on the same level and 1 when all targeted and not targeted classes are unlearned. We aim to obtain a \textbf{small} average score.

\subsection{Results}
\label{sec:results}

\begin{table*}[!t]
 \caption{We compare our method (Lip) to six methods averaging across three classes for four selected datasets. We aim to minimize \textit{Avg. Target Class acc. AF} while maintaining \textit{Avg. Other Classes acc. AF} and other datasets at a similar level to that before forgetting (BF).  We bold the best results comparing only among the first four methods that are zero-shot methods for a fair comparison. }
    \centering
    \fontsize{20}{25}\selectfont
    \setlength{\tabcolsep}{6pt} 
    \resizebox{0.8\textwidth}{!}{\begin{tabular}{@{}clc|cc|cc|cc|cc|cc|cc|cc|c@{}} 
    \toprule 
    \multirow{2}{*}{\makecell[c]{\textbf{Method}}} & \multirow{2}{*}{\textbf{Dataset}} & \multirow{2}{*}{\makecell[c]{\textbf{Forgetting} \\ \textbf{Type}}} & \multicolumn{2}{c}{\makecell[c]{\textbf{Avg. Target} \\ \textbf{Class acc.}}} & \multicolumn{2}{c}{\makecell[c]{\textbf{Avg. Other} \\ \textbf{Classes acc.}}} &  \multicolumn{2}{c}{\makecell[c]{\textbf{Avg.} \\ \textbf{StanfordCars}}} & \multicolumn{2}{c}{\makecell[c]{\textbf{Avg.} \\ \textbf{StanfordDogs}}} & \multicolumn{2}{c}{\makecell[c]{\textbf{Avg.} \\ \textbf{Caltech101}}} & \multicolumn{2}{c}{\makecell[c]{\textbf{Avg.} \\ \textbf{OxfordFlowers}}}  & \multirow{1}{*}{\makecell[c]{\textbf{Avg. Score} ($\downarrow$)}} \\\\
    \cline{4-16}
    & & & \textbf{BF} & \textbf{AF} & \textbf{BF} & \textbf{AF} & \textbf{BF} & \textbf{AF} & \textbf{BF} & \textbf{AF} & \textbf{BF} & \textbf{AF} & \textbf{BF} & \textbf{AF} \\
    \midrule
    Lip (Ours) & StanfordCars & ZS & 0.397 & 0.056 & 0.558 & 0.551 & - & - & 0.517 & 0.513 &  0.857 & 0.86  & 0.661 & 0.653 & \textbf{0.034} \\
    Emb & StanfordCars & ZS & 0.397 & 0.087 & 0.558 & 0.536 & - & - & 0.517 & 0.51 &  0.857 & 0.85  & 0.661 & 0.649 & 0.06 \\
    Amns\cite{graves_2020_amnesiac} & StanfordCars & ZS & 0.397 & 0.357 & 0.558 & 0.498 & - & - & 0.517 & 0.505 &  0.857 & 0.863  & 0.661 & 0.653 & 0.208 \\
    EMMN\cite{chundawat_2023_zeroshot} & StanfordCars & ZS & 0.397 & 0.0 & 0.558 & 0.054 & - & - & 0.517 & 0.043 &  0.857 & 0.424  & 0.661 & 0.069 & 0.644 \\
    \hdashline
    ULip\cite{foster_2024_zeroshot} & StanfordCars & semi ZS & 0.397 & 0.127 & 0.558 & 0.457 & - & - & 0.517 & 0.502 &  0.857 & 0.848  & 0.661 & 0.639 & 0.115 \\
    AmnsRetain\cite{graves_2020_amnesiac} & StanfordCars & not ZS & 0.397 & 0.04 & 0.558 & 0.711 & - & - & 0.517 & 0.509 &  0.857 & 0.881  & 0.661 & 0.622 & 0.035 \\
    Salun\cite{fan2023salun} & StanfordCars & not ZS & 0.397 & 0.063 & 0.558 & 0.712 & - & - & 0.517 & 0.491 &  0.857 & 0.862  & 0.661 & 0.574 & 0.068 \\
    \midrule
    Lip & StanfordDogs & ZS & 0.593 & 0.048 & 0.516 & 0.516 & 0.558 & 0.558 & - & - &  0.857 & 0.866  & 0.661 & 0.655 & \textbf{0.018} \\
    Emb & StanfordDogs & ZS & 0.593 & 0.261 & 0.516 & 0.479 & 0.558 & 0.554 & - & - &  0.857 & 0.836  & 0.661 & 0.621 & 0.121 \\
    Amns & StanfordDogs & ZS & 0.593 & 0.327 & 0.516 & 0.465 & 0.558 & 0.556 & - & - &  0.857 & 0.848  & 0.661 & 0.643 & 0.138 \\
    EMMN & StanfordDogs & ZS & 0.593 & 0.0 & 0.516 & 0.053 & 0.558 & 0.107 & - & - &  0.857 & 0.493  & 0.661 & 0.107 & 0.594 \\
    \hdashline
    ULip & StanfordDogs & semi ZS & 0.593 & 0.429 & 0.516 & 0.47 & 0.558 & 0.539 & - & - &  0.857 & 0.842  & 0.661 & 0.641 & 0.179 \\
    AmnsRetain & StanfordDogs & not ZS & 0.593 & 0.044 & 0.516 & 0.663 & 0.558 & 0.521 & - & - &  0.857 & 0.838  & 0.661 & 0.61 & 0.048 \\
    Salun & StanfordDogs & not ZS & 0.593 & 0.043 & 0.516 & 0.661 & 0.558 & 0.502 & - & - &  0.857 & 0.835  & 0.661 & 0.602 & 0.057 \\
    \midrule
    Lip & Caltech101 & ZS & 0.839 & 0.081 & 0.857 & 0.865 & 0.558 & 0.557 & 0.517 & 0.52 &  - & -  & 0.661 & 0.657 & \textbf{0.021} \\
    Emb & Caltech101 & ZS & 0.839 & 0.131 & 0.857 & 0.83 & 0.558 & 0.546 & 0.517 & 0.501 &  - & -  & 0.661 & 0.618 & 0.061 \\
    Amns & Caltech101 & ZS & 0.838 & 0.33 & 0.857 & 0.834 & 0.558 & 0.553 & 0.517 & 0.502 &  - & -  & 0.661 & 0.627 & 0.102 \\
    EMMN & Caltech101 & ZS & 0.839 & 0.0 & 0.857 & 0.397 & 0.558 & 0.097 & 0.517 & 0.081 &  - & -  & 0.661 & 0.13 & 0.602 \\
    \hdashline
    ULip & Caltech101 & semi ZS & 0.839 & 0.666 & 0.857 & 0.854 & 0.558 & 0.56 & 0.517 & 0.509 &  - & -  & 0.661 & 0.652 & 0.165 \\
    AmnsRetain & Caltech101 & not ZS & 0.839 & 0.0 & 0.857 & 0.925 & 0.558 & 0.526 & 0.517 & 0.505 &  - & -  & 0.661 & 0.636 & 0.023 \\
    Salun & Caltech101 & not ZS & 0.839 & 0.0 & 0.857 & 0.924 & 0.558 & 0.528 & 0.517 & 0.502 &  - & -  & 0.661 & 0.642 & 0.022 \\
    \midrule
    Lip & OxfordFlowers & ZS & 0.848 & 0.0 & 0.659 & 0.645 & 0.558 & 0.557 & 0.517 & 0.51 &  0.857 & 0.868  & - & - & \textbf{0.008} \\
    Emb & OxfordFlowers & ZS & 0.848 & 0.442 & 0.659 & 0.625 & 0.558 & 0.553 & 0.517 & 0.505 &  0.857 & 0.85  & - & - & 0.122 \\
    Amns & OxfordFlowers & ZS & 0.848 & 0.388 & 0.659 & 0.592 & 0.558 & 0.54 & 0.517 & 0.487 &  0.857 & 0.835  & - & - & 0.135 \\
    EMMN & OxfordFlowers & ZS & 0.848 & 0.0 & 0.659 & 0.121 & 0.558 & 0.121 & 0.517 & 0.112 &  0.857 & 0.676  & - & - & 0.519 \\
    \hdashline
    ULip & OxfordFlowers & semi ZS & 0.848 & 0.691 & 0.659 & 0.59 & 0.558 & 0.549 & 0.517 & 0.488 &  0.857 & 0.845  & - & - & 0.201 \\
    AmnsRetain & OxfordFlowers & not ZS & 0.848 & 0.059 & 0.659 & 0.922 & 0.558 & 0.553 & 0.517 & 0.51 &  0.857 & 0.866  & - & - & 0.018 \\
    Salun & OxfordFlowers & not ZS & 0.848 & 0.059 & 0.659 & 0.924 & 0.558 & 0.534 & 0.517 & 0.503 &  0.857 & 0.857  & - & - & 0.028 \\
    \bottomrule
    \end{tabular}}
   
    \label{table:forget_synth_aggreg}
\end{table*}

\begin{figure}[!b]
\centering
\includegraphics[width=1.\linewidth]{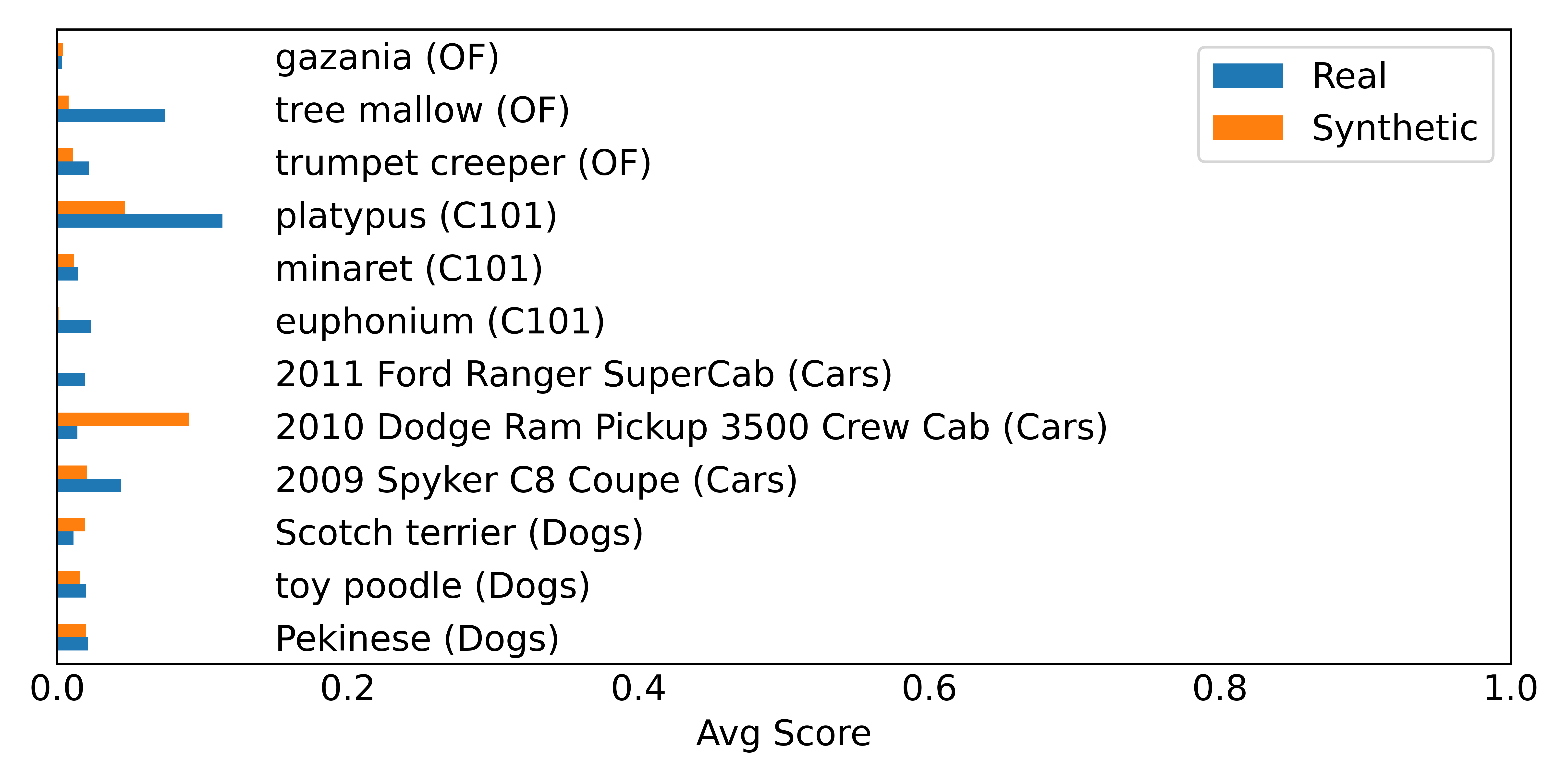}
\caption{Comparing the average scores of unlearning with Lip method using Synthetic vs Real data.}
\label{fig:synth_real_lip}
\end{figure}

\paragraph{Comparison across different methods}

In Tab. \ref{table:forget_synth_aggreg} we present the results of different forgetting methods averaging across three classes on four selected datasets. Full results can be found in Appendix \ref{app:rn_results}. \textit{Method} column refers to the method used in the experiments. \textit{Forgetting Type} 
refers to whether the method is zero-shot (ZS), indicating that no real data were used; semi zero-shot (semi ZS), where real data were used for forgetting; and not zero-shot (not ZS), where both retain and forget data were real. \textit{Dataset} column specifies the dataset from which the class to be forgotten was selected, while the \textit{Avg. Target Class acc.} denotes the average accuracy of the target class before (BF) and after (AF) forgetting. \textit{Avg. Other Classes acc.} indicates the average accuracy on the other classes in the dataset, excluding the class to be forgotten. The eight columns after that display the results on the remaining datasets reported both before and after forgetting. Finally, \textit{Avg. Score} is the aggregated metrics described previously. We observe that our forgetting procedure, referred to as \textit{Lip} has proven successful as indicated by a notable decrease in accuracy for the targeted classes, often approaching zero. Conversely, the accuracy for the remaining classes and other datasets remain largely unaffected after applying the forgetting procedure. This, indeed, results in the lowest average score across all examined methods. In comparison, the L2 embedding loss referred to as \textit{Emb} appears to be more aggressive than \textit{Lip}, not only erasing target knowledge but also impacting knowledge about other classes. We also tested L1 regularization loss that showed even more aggressive behaviour towards not targeted classes. We believe that the reason is that by enforcing a Lipschitz constraint we ensure that the output embeddings do not change too much in response to a small perturbation in the input (\textbf{ratio $\Delta$ output / $\Delta$ input}). This is particularly useful in unlearning as it prevents the model from drastically modifying the embeddings for non-targeted classes, hence preserving knowledge about non-target classes while only modifying the targeted ones. On the contrary, L1 and L2 regularization methods focus on penalizing the \textbf{magnitude} of the embeddings - L1 for sparsity, L2 for small embeddings. Thus, they do not inherently control how the output embeddings change to an input change. Thus applying L1 or L2 might lead to less stable unlearning, affecting non-targeted classes as we saw empirically. \\
On the other hand \textit{Amns} method has less forgetting power often resulting in not enough drop in accuracy of the target class and at the same time when forgetting was relatively successful results in over-forgetting on not targeted classes. The \textit{AmnsRetain} approach, which utilizes real data for both retention and forgetting, although not directly comparable to our method being not zero-shot, enables CLIP to forget the target class. We also observe that the accuracy on \textit{Avg. Other Classes acc. AF} is often much higher than \textit{BF} because of the classes to retain used for fine-tuning the model to regain the knowledge lost during forgetting. However, we note that datasets to which the forget class does not belong, and whose data was not used for retention, perform less effectively compared to our method. This demonstrates how large models like CLIP where we do not have information about the training data and classes suffer from drop in the accuracy on classes not included in the retain data. Therefore, our method not only competes in forgetting without using any real data and any retention data but also surpasses \textit{AmnsRetain} in terms of maintaining accuracy on other datasets. Similar conclusions to \textit{AmnsRetain} can be drawn for \textit{SalUn}. The \textit{EMMN} method, while often facilitating forgetting of the target class experiences significant decrease in accuracy, both on the not targeted classes of the dataset from which the forget class was picked and on other datasets. Finally, \textit{ULip} is not only not powerful enough to forget the target class but it also destroys knowledge not related to the target class resulting in a substantial drop on other classes of both related and unrelated datasets to the forget class. We attribute this phenomenon to asymmetric forgetting where attempting to erase knowledge in only one encoder disrupts the connection between the two encoders and consequently affects the projection in the shared embedding space. 

\paragraph{Comparison of our method with real and synthetic data}
We compare our method when using real and synthetic data on three classes for four different datasets in Fig. \ref{fig:synth_real_lip} with full details in Appendix \ref{app:synth_real_rn50}. We see that forgetting yields similar average scores for synthetic and real data. 

\paragraph{Multiple classes unlearning}
For experiments on unlearning multiple classes please refer to the Appendix \ref{app:err_mcls}.

\paragraph{Identity unlearning}
In line with our motivation of the right to be forgotten, we assess if our unlearning method enables identity unlearning using PinsFaces \cite{pinsfaces} dataset that contains 105 celebrity faces. These results are presented in the Appendix \ref{app:face_unlearn}. 

\subsection{Verification of Forgetting Success}

The accuracy achieved on synthetic forget data should serve as a measure of how effectively the model has forgotten a class. We find that for this indicator to be consistent the probability of the predicted class on synthetic samples need to be close to the probability of the real samples, otherwise there might be some discrepancy. We discuss this further in Appendix \ref{app:forg_success}.

\subsection{Predictions Before and After Forgetting on the Target Class}
\begin{figure}[!t]
\centering
\includegraphics[trim={5cm 1cm 5cm 1cm}, width=0.21\textwidth]{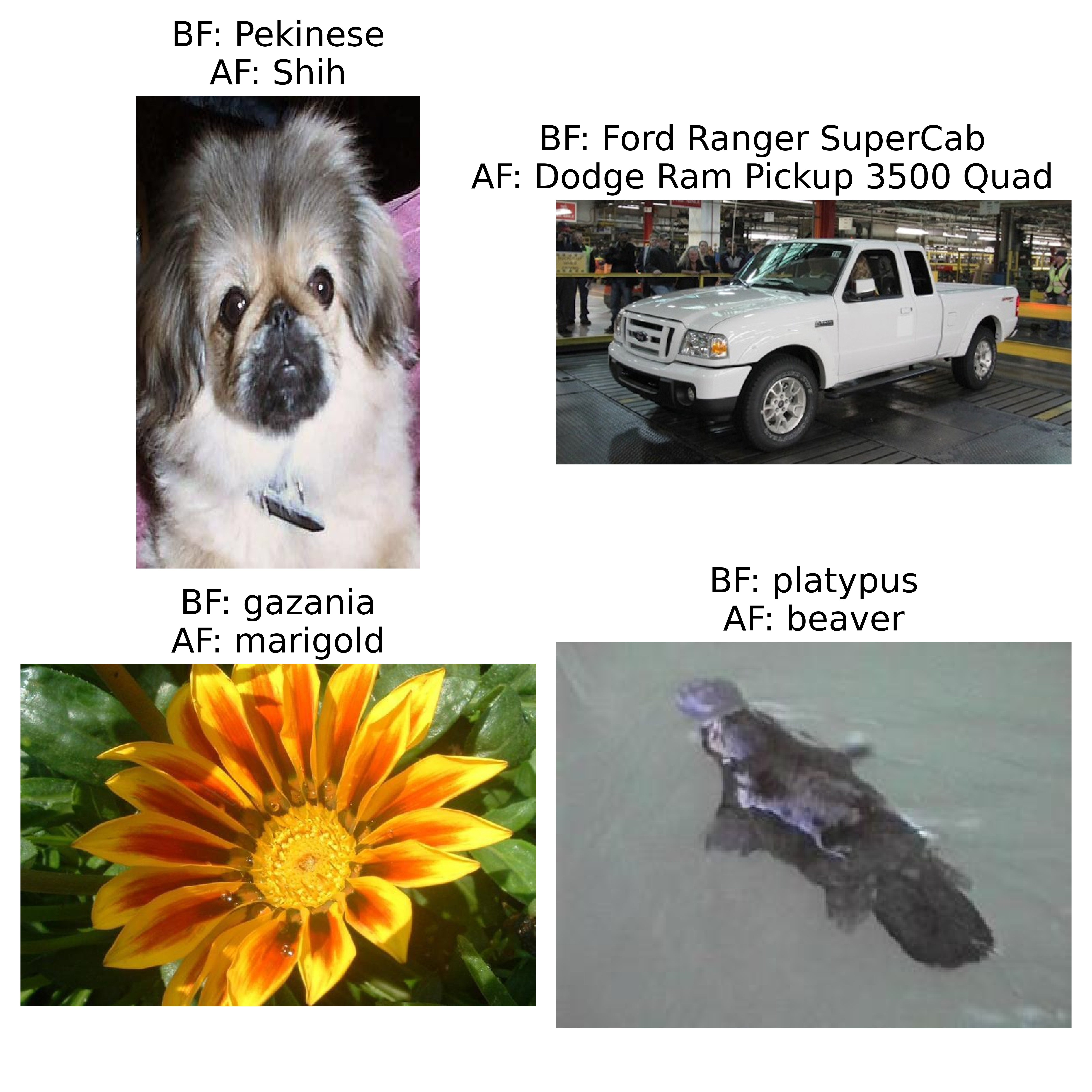}
\caption{Predictions of the model before (BF) and after forgetting (AF) with the prediction BF representing the target class to forget. }
\label{fig:cls_bf_af}
\end{figure}

In the Fig. \ref{fig:cls_bf_af} we present examples of the model's predictions before (BF) and after forgetting (AF). It's evident that post-forgetting, the model still predicts classes that closely resemble the correct ones, indicating that its general understanding of similar classes remains intact. This suggests that our method effectively targets specific knowledge of the model to remove detailed knowledge about the target class while preserving broader knowledge.

\begin{table}[!t]
\centering
\caption{Retrieval results showing precision@k for k of 1, 5, 10.}
\fontsize{7}{9}\selectfont 
\setlength{\tabcolsep}{3pt} 
\resizebox{0.45\textwidth}{!}{\begin{tabular}{@{}llccc@{}}
\toprule 
\textbf{Retrieval Type} & \textbf{Model} & \textbf{Precision@1 ($\downarrow$)} & \textbf{Precision@5 ($\downarrow$)} & \textbf{Precision@10 ($\downarrow$)} \\
\midrule
IfT & CLIP original Avg. & 0.833 & 0.683 & 0.583 \\
\midrule
IfT & CLIP forget Avg. & \textbf{0.08} & \textbf{0.23} & \textbf{0.191} \\
\midrule
\midrule
IfI & CLIP original Avg. & 0.417 & \textbf{0.367} & \textbf{0.317} \\
\midrule
IfI & CLIP forget Avg. & \textbf{0.333} & \textbf{0.367} & 0.325 \\
\bottomrule
\end{tabular}}
\label{table:imagetext_retr_IfT_IfI_rn50_main}
\end{table}

\begin{table*}[!t]

\caption{Different ablations. \textit{Original}: results with our method (Lip) from Tab.\ref{table:forget_synth_aggreg}. \textit{NoTextLoss}: ULip forgetting on synthetic data. \textit{CuPLGen}: synthetic samples generated with CuPL templates.  \textit{EvalTemplChange}: evaluation with different templates. }
    \centering
    \fontsize{20}{25}\selectfont
    \setlength{\tabcolsep}{4pt} 
    \resizebox{0.75\textwidth}{!}{\begin{tabular}{@{}lc|cc|cc|cc|cc|cc|cc|cc|c@{}} 
    \toprule 
    \multirow{2}{*}{\makecell[c]{\textbf{Ablation} \\ \textbf{Type}}} & \multirow{2}{*}{\makecell[c]{\textbf{Method}}} & \multicolumn{2}{c}{\makecell[c]{\textbf{Avg. Target} \\ \textbf{Class acc.}}} & \multicolumn{2}{c}{\makecell[c]{\textbf{Avg. Other} \\ \textbf{Classes acc.}}} & \multicolumn{2}{c}{\makecell[c]{\textbf{Avg.} \\ \textbf{StanfordCars}}} & \multicolumn{2}{c}{\makecell[c]{\textbf{Avg.} \\ \textbf{StanfordDogs}}} & \multicolumn{2}{c}{\makecell[c]{\textbf{Avg.} \\ \textbf{Caltech101}}} & \multicolumn{2}{c}{\makecell[c]{\textbf{Avg.} \\ \textbf{OxfordFlowers}}} & \multirow{1}{*}{\makecell[c]{\textbf{Avg. Score} ($\downarrow$)}} \\
    \cline{3-15}
    & & \textbf{BF} & \textbf{AF} & \textbf{BF} & \textbf{AF} & \textbf{BF} & \textbf{AF} & \textbf{BF} & \textbf{AF} & \textbf{BF} & \textbf{AF} & \textbf{BF} & \textbf{AF} \\
    \midrule
    Original & Lip & 0.669 & 0.046 & 0.648 & 0.644 & 0.558 & 0.557 & 0.517 & 0.514 &  0.857 & 0.865  & 0.661 & 0.655 & \textbf{0.018} \\
    \midrule
    NoTextLoss & ULip & 0.669 & 0.606 & 0.648 & 0.641 & 0.558 & 0.557 & 0.517 & 0.511 &  0.857 & 0.854  & 0.661 & 0.651 & 0.189 \\ 
    \midrule
    CuPLGen & Lip & 0.669 & 0.038 & 0.648 & 0.609 & 0.558 & 0.535 & 0.517 & 0.494 &  0.857 & 0.838  & 0.661 & 0.625 & 0.055 \\
    \midrule
    EvalTemplChange & Lip & 0.522 & 0.133 & 0.544 & 0.538 & 0.492 & 0.494 & 0.412 & 0.414 &  0.81 & 0.801  & 0.518 & 0.519 & 0.068 \\
    
    \bottomrule
    \end{tabular}}
   
    \label{table:aggregated_ablations}
\end{table*}

\subsection{Additional Tasks}
 
We evaluate our method for the retrieval task in addition to classification. The retrieval task involves text retrieval from the image input, image retrieval from the text input, and image retrieval from the image input. As classification can be viewed as text retrieval from an image, we present aggregated results for the other two retrieval tasks in Tab. \ref{table:imagetext_retr_IfT_IfI_rn50_main}, with full results provided in Appendix \ref{app:addtasks}.

\paragraph{Image Retrieval from Text Input (IfT)}
We create a database from 4 datasets and perform image retrieval task  given a text input. We evaluate on precision@k metric measuring the proportion of retrieved items that are relevant among top K retrieved items. This indicates the accuracy of the retrieved results. We perform our experiments with k of 1, 5 and 10. Note that the lower the precision@k the better. We see in Tab. \ref{table:imagetext_retr_IfT_IfI_rn50_main} that the model is most of the times unable to retrieve images from input text. 

\paragraph{Image Retrieval from Image Input (IfI)}
Similarly to above, but now we test image-image retrieval. We observe in Tab. \ref{table:imagetext_retr_IfT_IfI_rn50_main} that image representation for the forget objects is mainly untouched and the model is able to find also forget classes. These results indicate that forgetting is achieved breaking the multimodal link but unimodal information still remains in the model. This was surprising and we ask whether our forgetting is successful given these results? Therefore, we checked whether the original CLIP is able to retrieve images from image input for classes the model is unable to classify, i.e. has classification accuracy of 0. In Appendix \ref{app:addtasks} we show that the model can identify similar features and shapes of objects without knowing the textual class. Thus, we conclude that for class forgetting, breaking the text-image association is sufficient.

\section{Ablations}
In this section we present ablations to understand the sensitivity of our method to changes in a) prompts to generate synthetic samples, b) evaluation templates and c) unlearning settings. For extra ablations and granular results please refer to Appendix \ref{app:extra_abls}. 

\subsection{Textual Loss Ablation}
In Tab. \ref{table:aggregated_ablations} we compare ULip (\textit{NoTextLoss} ablation type) method and our Lip method (\textit{Original} ablation type) using synthetic data. The only difference between these methods is the inclusion of an additional textual loss in the Lip method. The results demonstrate the critical importance of incorporating both visual and textual losses for effective forgetting in CLIP, as ULip forgetting with synthetic data proves to be highly ineffective.

\subsection{Synthetic Images Generation with CuPL Template}
We test how the text templates of the generated samples affect performance. We generate synthetic samples using templates from CuPL\footnote{https://github.com/sarahpratt/CuPL/tree/main/all\_prompts}, e.g. for the "Pekinese" class one example is \textit{"The image is of a small, brown and white Pekinese dog with long, flowing fur."}. We generate 64 synthetic samples and for each generated sample a random description of the class from CuPL is used. CuPL descriptions involve not only the class itself but also features of the class containing thus more information additionally to the class name. The results are shown in Tab. \ref{table:aggregated_ablations} in \textit{CuPLGen} ablation type and more granular results in Appendix \ref{app:extra_abls}. We see that by changing the template of the synthetic samples generation the forgetting is still successful in breaking the image-text association for the class. However, because of additional features in the template that might be shared among other classes such as \textit{"flowing fur"} the remaining accuracy slightly decreases. Therefore, forgetting can be sensitive to the template used to generate the synthetic samples. Also note that the evaluation is still performed using the standard template.

\subsection{Variation of Templates for Evaluation}
In the following experiments we test the sensitivity of the model after forgetting to the change in the evaluation template. We use the synthetic samples generated with a standard template but evaluate using three different templates: \textit{"We can see a \{class\} in this image"}, \textit{"This is a representation of \{class\}"}, \textit{"There is evidence of a \{class\} in the picture"}. This shows how sensitive the model's evaluation is to the change in template after forgetting. Note that because the evaluation template changed, so did the zero-shot classification accuracy before forgetting on CLIP. The results are shown in Tab. \ref{table:aggregated_ablations} in \textit{EvalTemplChange} ablation type where we observe that even after changing the evaluation template forgetting is still valid across the classes we have forgotten. More granular results can be seen in Appendix \ref{app:extra_abls}.

\section{Limitations}
\label{app:limit}
One limitation is that it is hard to assess how well the model will perform on other classes after unlearning. Looking at \textit{2012 Chevrolet Avalanche Crew Cab} class in Tab. \ref{table:forget_other} in the Appendix, even if forgetting is quite successful, 6\% of accuracy is lost on other classes of \textit{StanfordCars}. Note that knowledge about classes not related to cars remained fairly close to that before forgetting. Our iterative procedure can help control this trade-off between unlearning and retaining knowledge. Another limitation is that forgetting certain classes is harder and additional tuning of forgetting aggressiveness parameters and the synthetic data generation threshold might be required.

\section{Conclusions}
In this work we have successfully achieved class forgetting without losing knowledge on other classes in the multimodal setting of CLIP. Our experiments were conducted on four standard datasets, demonstrating that forgetting can be achieved based solely on the textual class names by generating synthetic samples of the class, without dependence on real data, thus achieving true zero-shot forgetting. Our forgetting process is iterative where we increase the number of layers to update and the strength of perturbations based on the reduction in accuracy of synthetic training data. 

\paragraph{Acknowledgements}
We'd like to gratefully acknowledge Microsoft’s compute support through Microsoft’s Accelerating Foundation Models Research grant and the support from University of Bath for the studentship. 


{\small
\bibliographystyle{ieee_fullname}
\bibliography{egbib}
}

\newpage
\clearpage

\iftoggle{papermain}{
\appendix

\onecolumn
{\huge \textbf{Appendix Table of Contents}}

\begin{itemize}
    \item \textbf{ResNet Full Results: we include full results for ResNet model. } \hfill \pageref{app:rn_results}
    \item \textbf{Additional Ablations: includes full results on ablations presented in the main paper and perturbations to text encoder ablation. } \hfill \pageref{app:extra_abls}
    \item \textbf{Lip Unlearning Real vs Synthetic: contains tabular results comparing unlearning with synthetic and real data with our method. } \hfill \pageref{app:synth_real_rn50}
    \item \textbf{Forgetting on Multiple Classes and Error Analysis: results on unlearning with multiple classes with ResNet model. } \hfill \pageref{app:err_mcls}
    \item \textbf{Unlearning Faces Full Results. } \hfill \pageref{app:face_unlearn}
    \item \textbf{Synthetic Images Visualization from ResNet model. } \hfill \pageref{app:synth_vis}
    \item \textbf{ViT Results: results for ViT-B/16 unlearning including multiple class unlearning and comparison of unlearnig of real vs synthetic data. } \hfill \pageref{app:vit_results}
    \item \textbf{Forgetting Algorithm. } \hfill \pageref{app:forg_algo}
    \item \textbf{Verification of Forgetting Success and Data
 Generation Threshold: further discussion and analysis on forgetting success verification and data generation threshold.} \hfill \pageref{app:forg_success}
    \item \textbf{Additional Tasks: retrieval of image from text and image from image after unlearning.  } \hfill \pageref{app:addtasks}
    \item \textbf{Additional Figures and Implementation Details: implementations details and additional figures. } \hfill \pageref{app:add_impl}
\end{itemize}

\onecolumn
\section{ResNet Full Results}
\label{app:rn_results}
\addcontentsline{toc}{section}{ResNet Full Results} 

\begin{table*}[!h]
 \caption{Forgetting results with ResNet visual encoder. We compare our methods with five others on three classes for four selected datasets. We bold the best results comparing only among the first four methods that are zero-shot methods for a fair comparison.}
    \centering
    \fontsize{20}{25}\selectfont
    \setlength{\tabcolsep}{4pt} 
    \resizebox{0.75\textwidth}{!}{\begin{tabular}{@{}clc|cc|cc|cc|cc|cc|cc|cc|c@{}} 
    \toprule 
    \multirow{2}{*}{\makecell[c]{\textbf{Method}}} & \multirow{2}{*}{\textbf{Dataset}} & \multirow{2}{*}{\textbf{Class name}} & \multicolumn{2}{c}{\makecell[c]{\textbf{Target} \\ \textbf{Class acc.}}} & \multicolumn{2}{c}{\makecell[c]{\textbf{Other} \\ \textbf{Classes acc.}}} & \multicolumn{2}{c}{\makecell[c]{\textbf{Target} \\ \textbf{Class acc.}}} & \multicolumn{2}{c}{\textbf{StanfordCars}} & \multicolumn{2}{c}{\textbf{StanfordDogs}} & \multicolumn{2}{c}{\textbf{Caltech101}} & \multicolumn{2}{c}{\textbf{OxfordFlowers}} & \multirow{1}{*}{\makecell[c]{\textbf{Avg. Score} ($\downarrow$)}} \\
    \cline{4-18}
    & & & \textbf{BF} & \textbf{AF} & \textbf{BF} & \textbf{AF} & {\makecell[c]{\textbf{Synt.} \\ \textbf{train}}} & {\makecell[c]{\textbf{Real} \\ \textbf{valid.}}} & \textbf{BF} & \textbf{AF} & \textbf{BF} & \textbf{AF} & \textbf{BF} & \textbf{AF} & \textbf{BF} & \textbf{AF} & \\
    \midrule
    
    Lip & StanfordDogs & Pekinese & 0.705 & 0.066 & 0.515 & 0.514 & 0.062 & 0.0  & 0.558 & 0.559 & - & - &  0.857 & 0.867  & 0.661 & 0.658 & \textbf{0.02} \\
    Lip & StanfordDogs & toy poodle & 0.574 & 0.033 & 0.516 & 0.518 & 0.031 & 0.0  & 0.558 & 0.559 & - & - &  0.857 & 0.867  & 0.661 & 0.647 & \textbf{0.016} \\
    Lip & StanfordDogs & Scotch terrier & 0.5 & 0.047 & 0.517 & 0.516 & 0.047 & 0.083  & 0.558 & 0.557 & - & - &  0.857 & 0.865  & 0.661 & 0.66 & \textbf{0.019} \\
    Lip & StanfordCars & 2009 Spyker C8 Coupe & 0.262 & 0.024 & 0.559 & 0.553 & 0.0 & 0.0  & - & - & 0.517 & 0.518 &  0.857 & 0.865  & 0.661 & 0.66 & \textbf{0.021} \\
    Lip & StanfordCars & 2010 Dodge Ram Pickup 3500 Crew Cab & 0.405 & 0.143 & 0.558 & 0.544 & 0.0 & 0.0  & - & - & 0.517 & 0.502 &  0.857 & 0.845  & 0.661 & 0.638 & 0.091 \\
    Lip & StanfordCars & 2011 Ford Ranger SuperCab & 0.524 & 0.0 & 0.558 & 0.555 & 0.0 & 0.0  & - & - & 0.517 & 0.52 &  0.857 & 0.869  & 0.661 & 0.661 & \textbf{0.001} \\
    Lip & Caltech101 & euphonium & 0.789 & 0.0 & 0.858 & 0.868 & 0.016 & 0.0  & 0.558 & 0.557 & 0.517 & 0.52 &  - & -  & 0.661 & 0.658 & \textbf{0.001} \\
    Lip & Caltech101 & minaret & 0.826 & 0.043 & 0.857 & 0.863 & 0.0 & 0.067  & 0.558 & 0.556 & 0.517 & 0.515 &  - & -  & 0.661 & 0.661 & \textbf{0.012} \\
    Lip & Caltech101 & platypus & 0.9 & 0.2 & 0.857 & 0.866 & 0.062 & 0.286  & 0.558 & 0.558 & 0.517 & 0.524 &  - & -  & 0.661 & 0.653 & 0.047 \\
    Lip & OxfordFlowers & gazania & 0.957 & 0.0 & 0.658 & 0.649 & 0.062 & 0.0  & 0.558 & 0.559 & 0.517 & 0.513 &  0.857 & 0.869  & - & - & \textbf{0.004} \\
    Lip & OxfordFlowers & tree mallow & 1.0 & 0.0 & 0.658 & 0.643 & 0.047 & 0.0  & 0.558 & 0.557 & 0.517 & 0.51 &  0.857 & 0.869  & - & - & \textbf{0.008} \\
    Lip & OxfordFlowers & trumpet creeper & 0.588 & 0.0 & 0.661 & 0.643 & 0.047 & 0.083  & 0.558 & 0.557 & 0.517 & 0.503 &  0.857 & 0.866  & - & - & \textbf{0.011} \\
    \midrule
 
    Emb & StanfordDogs & Pekinese & 0.705 & 0.361 & 0.515 & 0.484 & 0.0 & 0.318  & 0.558 & 0.559 & - & - &  0.857 & 0.84  & 0.661 & 0.633 & 0.127 \\
    Emb & StanfordDogs & toy poodle & 0.574 & 0.361 & 0.516 & 0.481 & 0.0 & 0.217  & 0.558 & 0.553 & - & - &  0.857 & 0.832  & 0.661 & 0.613 & 0.161 \\
    Emb & StanfordDogs & Scotch terrier & 0.5 & 0.062 & 0.517 & 0.472 & 0.031 & 0.083  & 0.558 & 0.551 & - & - &  0.857 & 0.837  & 0.661 & 0.617 & 0.063 \\
    Emb & StanfordCars & 2009 Spyker C8 Coupe & 0.262 & 0.024 & 0.559 & 0.529 & 0.0 & 0.0  & - & - & 0.517 & 0.508 &  0.857 & 0.841  & 0.661 & 0.639 & 0.043 \\
    Emb & StanfordCars & 2010 Dodge Ram Pickup 3500 Crew Cab & 0.405 & 0.119 & 0.558 & 0.542 & 0.047 & 0.0  & - & - & 0.517 & 0.512 &  0.857 & 0.857  & 0.661 & 0.654 & \textbf{0.069} \\
    Emb & StanfordCars & 2011 Ford Ranger SuperCab & 0.524 & 0.119 & 0.558 & 0.539 & 0.0 & 0.125  & - & - & 0.517 & 0.509 &  0.857 & 0.852  & 0.661 & 0.654 & 0.059 \\
    Emb & Caltech101 & euphonium & 0.789 & 0.263 & 0.858 & 0.833 & 0.0 & 0.385  & 0.558 & 0.548 & 0.517 & 0.506 &  - & -  & 0.661 & 0.616 & 0.093 \\
    Emb & Caltech101 & minaret & 0.826 & 0.13 & 0.857 & 0.827 & 0.0 & 0.133  & 0.558 & 0.54 & 0.517 & 0.507 &  - & -  & 0.661 & 0.639 & 0.055 \\
    Emb & Caltech101 & platypus & 0.9 & 0.0 & 0.857 & 0.829 & 0.047 & 0.0  & 0.558 & 0.549 & 0.517 & 0.49 &  - & -  & 0.661 & 0.597 & \textbf{0.039} \\
    Emb & OxfordFlowers & gazania & 0.957 & 0.739 & 0.658 & 0.632 & 0.0 & 0.875  & 0.558 & 0.551 & 0.517 & 0.503 &  0.857 & 0.849  & - & - & 0.172 \\
    Emb & OxfordFlowers & tree mallow & 1.0 & 0.353 & 0.658 & 0.612 & 0.094 & 0.583  & 0.558 & 0.554 & 0.517 & 0.504 &  0.857 & 0.849  & - & - & 0.093 \\
    Emb & OxfordFlowers & trumpet creeper & 0.588 & 0.235 & 0.661 & 0.632 & 0.0 & 0.167  & 0.558 & 0.555 & 0.517 & 0.508 &  0.857 & 0.853  & - & - & 0.094 \\

    \midrule
    
    Amns & StanfordDogs & Pekinese & 0.705 & 0.459 & 0.515 & 0.486 & 0.0 & 0.409  & 0.558 & 0.561 & - & - &  0.857 & 0.847  & 0.661 & 0.65 & 0.147 \\
    Amns & StanfordDogs & toy poodle & 0.574 & 0.492 & 0.516 & 0.423 & 0.016 & 0.261  & 0.558 & 0.55 & - & - &  0.857 & 0.839  & 0.661 & 0.628 & 0.225 \\
    Amns & StanfordDogs & Scotch terrier & 0.5 & 0.031 & 0.517 & 0.488 & 0.0 & 0.083  & 0.558 & 0.559 & - & - &  0.857 & 0.859  & 0.661 & 0.651 & 0.027 \\
    Amns & StanfordCars & 2009 Spyker C8 Coupe & 0.262 & 0.143 & 0.559 & 0.516 & 0.016 & 0.0  & - & - & 0.517 & 0.51 &  0.857 & 0.854  & 0.661 & 0.646 & 0.132 \\
    Amns & StanfordCars & 2010 Dodge Ram Pickup 3500 Crew Cab & 0.405 & 0.429 & 0.558 & 0.49 & 0.047 & 0.222  & - & - & 0.517 & 0.5 &  0.857 & 0.868  & 0.661 & 0.658 & 0.244 \\
    Amns & StanfordCars & 2011 Ford Ranger SuperCab & 0.524 & 0.5 & 0.558 & 0.489 & 0.078 & 0.375  & - & - & 0.517 & 0.507 &  0.857 & 0.868  & 0.661 & 0.656 & 0.221 \\
    Amns & Caltech101 & euphonium & 0.789 & 0.316 & 0.858 & 0.856 & 0.016 & 0.385  & 0.558 & 0.557 & 0.517 & 0.519 &  - & -  & 0.661 & 0.655 & 0.082 \\
    Amns & Caltech101 & platypus & 0.9 & 0.5 & 0.857 & 0.832 & 0.0 & 0.571  & 0.558 & 0.555 & 0.517 & 0.495 &  - & -  & 0.661 & 0.634 & 0.134 \\
    Amns & Caltech101 & minaret & 0.826 & 0.174 & 0.857 & 0.813 & 0.031 & 0.267  & 0.558 & 0.546 & 0.517 & 0.493 &  - & -  & 0.661 & 0.591 & 0.087 \\
    Amns & OxfordFlowers & gazania & 0.957 & 0.87 & 0.658 & 0.595 & 0.391 & 0.875  & 0.558 & 0.557 & 0.517 & 0.489 &  0.857 & 0.834  & - & - & 0.217 \\
    Amns & OxfordFlowers & tree mallow & 1.0 & 0.0 & 0.658 & 0.598 & 0.094 & 0.0  & 0.558 & 0.511 & 0.517 & 0.476 &  0.857 & 0.843  & - & - & 0.054 \\
    Amns & OxfordFlowers & trumpet creeper & 0.588 & 0.294 & 0.661 & 0.584 & 0.0 & 0.25  & 0.558 & 0.554 & 0.517 & 0.494 &  0.857 & 0.828  & - & - & 0.14 \\
    
    \midrule
   
    EMMN & StanfordDogs & Pekinese & 0.705 & 0.0 & 0.515 & 0.073 & - & -  & 0.558 & 0.102 & - & - &  0.857 & 0.574  & 0.661 & 0.129 & 0.562 \\
    EMMN & StanfordDogs & toy poodle & 0.574 & 0.0 & 0.516 & 0.041 & - & -  & 0.558 & 0.118 & - & - &  0.857 & 0.622  & 0.661 & 0.13 & 0.557 \\
    EMMN & StanfordDogs & Scotch terrier & 0.5 & 0.0 & 0.517 & 0.046 & - & -  & 0.558 & 0.1 & - & - &  0.857 & 0.284  & 0.661 & 0.061 & 0.661 \\
    EMMN & StanfordCars & 2009 Spyker C8 Coupe & 0.262 & 0.0 & 0.559 & 0.057 & - & -  & - & - & 0.517 & 0.055 &  0.857 & 0.491  & 0.661 & 0.07 & 0.623 \\
    EMMN & StanfordCars & 2010 Dodge Ram Pickup 3500 Crew Cab & 0.405 & 0.0 & 0.558 & 0.049 & - & -  & - & - & 0.517 & 0.042 &  0.857 & 0.435  & 0.661 & 0.068 & 0.644 \\
    EMMN & StanfordCars & 2011 Ford Ranger SuperCab & 0.524 & 0.0 & 0.558 & 0.058 & - & -  & - & - & 0.517 & 0.032 &  0.857 & 0.348  & 0.661 & 0.07 & 0.665 \\
    EMMN & Caltech101 & euphonium & 0.789 & 0.0 & 0.858 & 0.442 & - & -  & 0.558 & 0.107 & 0.517 & 0.083 &  - & -  & 0.661 & 0.173 & 0.574 \\
    EMMN & Caltech101 & minaret & 0.826 & 0.0 & 0.857 & 0.34 & - & -  & 0.558 & 0.106 & 0.517 & 0.093 &  - & -  & 0.661 & 0.119 & 0.611 \\
    EMMN & Caltech101 & platypus & 0.9 & 0.0 & 0.857 & 0.411 & - & -  & 0.558 & 0.078 & 0.517 & 0.066 &  - & -  & 0.661 & 0.098 & 0.621 \\
    EMMN & OxfordFlowers & gazania & 0.957 & 0.0 & 0.658 & 0.107 & - & -  & 0.558 & 0.11 & 0.517 & 0.104 &  0.857 & 0.643  & - & - & 0.538 \\
    EMMN & OxfordFlowers & tree mallow & 1.0 & 0.0 & 0.658 & 0.115 & - & -  & 0.558 & 0.136 & 0.517 & 0.089 &  0.857 & 0.66  & - & - & 0.528 \\
    EMMN & OxfordFlowers & trumpet creeper & 0.588 & 0.0 & 0.661 & 0.14 & - & -  & 0.558 & 0.116 & 0.517 & 0.142 &  0.857 & 0.725  & - & - & 0.492 \\
    \midrule
    
    AmnsRetain & StanfordDogs & Pekinese & 0.705 & 0.049 & 0.515 & 0.667 & - & - & 0.558 & 0.531 & - & - &  0.857 & 0.835  & 0.661 & 0.61 & 0.044 \\
    AmnsRetain & StanfordDogs & toy poodle & 0.574 & 0.082 & 0.516 & 0.663 & - & - & 0.558 & 0.521 & - & - &  0.857 & 0.831  & 0.661 & 0.605 & 0.065 \\
    AmnsRetain & StanfordDogs & Scotch terrier & 0.5 & 0.0 & 0.517 & 0.659 & - & - & 0.558 & 0.511 & - & - &  0.857 & 0.847  & 0.661 & 0.616 & 0.033 \\
    AmnsRetain & StanfordCars & 2009 Spyker C8 Coupe & 0.262 & 0.071 & 0.559 & 0.719 & - & - & - & - & 0.517 & 0.513 &  0.857 & 0.886  & 0.661 & 0.623 & 0.067 \\
    AmnsRetain & StanfordCars & 2010 Dodge Ram Pickup 3500 Crew Cab & 0.405 & 0.0 & 0.558 & 0.712 & - & - & - & - & 0.517 & 0.504 &  0.857 & 0.878  & 0.661 & 0.62 & 0.017 \\
    AmnsRetain & StanfordCars & 2011 Ford Ranger SuperCab & 0.524 & 0.048 & 0.558 & 0.704 & - & - & - & - & 0.517 & 0.51 &  0.857 & 0.879  & 0.661 & 0.622 & 0.033 \\
    AmnsRetain & Caltech101 & euphonium & 0.789 & 0.0 & 0.858 & 0.926 & - & - & 0.558 & 0.537 & 0.517 & 0.506 &  - & -  & 0.661 & 0.624 & 0.022 \\
    AmnsRetain & Caltech101 & minaret & 0.826 & 0.0 & 0.857 & 0.927 & - & - & 0.558 & 0.516 & 0.517 & 0.501 &  - & -  & 0.661 & 0.63 & 0.03 \\
    AmnsRetain & Caltech101 & platypus & 0.9 & 0.0 & 0.857 & 0.923 & - & - & 0.558 & 0.526 & 0.517 & 0.508 &  - & -  & 0.661 & 0.654 & 0.017 \\
    AmnsRetain & OxfordFlowers & gazania & 0.957 & 0.0 & 0.658 & 0.931 & - & - & 0.558 & 0.554 & 0.517 & 0.512 &  0.857 & 0.865  & - & - & 0.004 \\
    AmnsRetain & OxfordFlowers & tree mallow & 1.0 & 0.0 & 0.658 & 0.931 & - & - & 0.558 & 0.552 & 0.517 & 0.503 &  0.857 & 0.872  & - & - & 0.008 \\
    AmnsRetain & OxfordFlowers & trumpet creeper & 0.588 & 0.176 & 0.661 & 0.903 & - & - & 0.558 & 0.553 & 0.517 & 0.515 &  0.857 & 0.861  & - & - & 0.062 \\
    \midrule

    Salun & StanfordDogs & Pekinese & 0.705 & 0.049 & 0.515 & 0.668 & - & - & 0.558 & 0.509 & - & - &  0.857 & 0.838  & 0.661 & 0.605 & 0.053 \\
    Salun & StanfordDogs & toy poodle & 0.574 & 0.066 & 0.516 & 0.654 & - & - & 0.558 & 0.502 & - & - &  0.857 & 0.841  & 0.661 & 0.614 & 0.061 \\
    Salun & StanfordDogs & Scotch terrier & 0.5 & 0.016 & 0.517 & 0.661 & - & - & 0.558 & 0.497 & - & - &  0.857 & 0.826  & 0.661 & 0.587 & 0.058 \\
    Salun & StanfordCars & 2009 Spyker C8 Coupe & 0.262 & 0.119 & 0.559 & 0.718 & - & - & - & - & 0.517 & 0.498 &  0.857 & 0.871  & 0.661 & 0.58 & 0.123 \\
    Salun & StanfordCars & 2010 Dodge Ram Pickup 3500 Crew Cab & 0.405 & 0.024 & 0.558 & 0.711 & - & - & - & - & 0.517 & 0.494 &  0.857 & 0.853  & 0.661 & 0.563 & 0.051 \\
    Salun & StanfordCars & 2011 Ford Ranger SuperCab & 0.524 & 0.048 & 0.558 & 0.707 & - & - & - & - & 0.517 & 0.481 &  0.857 & 0.862  & 0.661 & 0.58 & 0.056 \\
    Salun & Caltech101 & euphonium & 0.789 & 0.0 & 0.858 & 0.924 & - & - & 0.558 & 0.531 & 0.517 & 0.498 &  - & -  & 0.661 & 0.65 & 0.02 \\
    Salun & Caltech101 & minaret & 0.826 & 0.0 & 0.857 & 0.925 & - & - & 0.558 & 0.522 & 0.517 & 0.508 &  - & -  & 0.661 & 0.638 & 0.023 \\
    Salun & Caltech101 & platypus & 0.9 & 0.0 & 0.857 & 0.923 & - & - & 0.558 & 0.532 & 0.517 & 0.5 &  - & -  & 0.661 & 0.637 & 0.023 \\
    Salun & OxfordFlowers & gazania & 0.957 & 0.0 & 0.658 & 0.92 & - & - & 0.558 & 0.521 & 0.517 & 0.503 &  0.857 & 0.86  & - & - & 0.018 \\
    Salun & OxfordFlowers & tree mallow & 1.0 & 0.059 & 0.658 & 0.928 & - & - & 0.558 & 0.539 & 0.517 & 0.502 &  0.857 & 0.858  & - & - & 0.024 \\
    Salun & OxfordFlowers & trumpet creeper & 0.588 & 0.118 & 0.661 & 0.922 & - & - & 0.558 & 0.541 & 0.517 & 0.503 &  0.857 & 0.853  & - & - & 0.052 \\

    \midrule
    
    ULip & StanfordDogs & Pekinese & 0.705 & 0.705 & 0.515 & 0.506 & - & - & 0.558 & 0.563 & - & - &  0.857 & 0.863  & 0.661 & 0.656 & 0.205 \\
    ULip & StanfordDogs & toy poodle & 0.574 & 0.082 & 0.516 & 0.395 & - & - & 0.558 & 0.494 & - & - &  0.857 & 0.81  & 0.661 & 0.611 & 0.124 \\
    ULip & StanfordDogs & Scotch terrier & 0.5 & 0.5 & 0.517 & 0.509 & - & - & 0.558 & 0.561 & - & - &  0.857 & 0.853  & 0.661 & 0.657 & 0.205 \\
    ULip & StanfordCars & 2009 Spyker C8 Coupe & 0.262 & 0.119 & 0.559 & 0.389 & - & - & - & - & 0.517 & 0.492 &  0.857 & 0.825  & 0.661 & 0.63 & 0.178 \\
    ULip & StanfordCars & 2010 Dodge Ram Pickup 3500 Crew Cab & 0.405 & 0.071 & 0.558 & 0.497 & - & - & - & - & 0.517 & 0.51 &  0.857 & 0.859  & 0.661 & 0.643 & 0.065 \\
    ULip & StanfordCars & 2011 Ford Ranger SuperCab & 0.524 & 0.19 & 0.558 & 0.486 & - & - & - & - & 0.517 & 0.504 &  0.857 & 0.86  & 0.661 & 0.644 & 0.109 \\
    ULip & Caltech101 & euphonium & 0.789 & 0.316 & 0.858 & 0.849 & - & - & 0.558 & 0.556 & 0.517 & 0.5 &  - & -  & 0.661 & 0.64 & 0.095 \\
    ULip & Caltech101 & minaret & 0.826 & 0.783 & 0.857 & 0.855 & - & - & 0.558 & 0.564 & 0.517 & 0.514 &  - & -  & 0.661 & 0.659 & 0.192 \\
    ULip & Caltech101 & platypus & 0.9 & 0.9 & 0.857 & 0.858 & - & - & 0.558 & 0.561 & 0.517 & 0.513 &  - & -  & 0.661 & 0.657 & 0.203 \\
    ULip & OxfordFlowers & gazania & 0.957 & 0.957 & 0.658 & 0.64 & - & - & 0.558 & 0.559 & 0.517 & 0.51 &  0.857 & 0.856  & - & - & 0.208 \\
    ULip & OxfordFlowers & tree mallow & 1.0 & 1.0 & 0.658 & 0.646 & - & - & 0.558 & 0.56 & 0.517 & 0.51 &  0.857 & 0.86  & - & - & 0.207 \\
    ULip & OxfordFlowers & trumpet creeper & 0.588 & 0.118 & 0.661 & 0.483 & - & - & 0.558 & 0.528 & 0.517 & 0.445 &  0.857 & 0.82  & - & - & 0.141 \\
    \bottomrule
    \end{tabular}}
   
    \label{table:full_forget_synth}
\end{table*}

\clearpage
\twocolumn

\section{Additional Ablations}
\label{app:extra_abls}

\subsection{Full Results Synthetic Sampels CuPL Templates }
Tab. \ref{table:forget_synth_cupl} contains more granular results of forgetting on synthetic samples generated with randomly selected CuPL templates. 

\subsection{Full Results Evaluation with Different Template }
Tab. \ref{table:forget_synth_changetemplate} contains more granular results of forgetting evaluation with different templates.

\subsection{Full Results Forgetting of ULip with Synthetic Samples }
Tab. \ref{table:ulip_synth} contains more granular results of forgetting with ULip with synthetic samples.

\subsection{Varying All the Parameters}
When all parameters are allowed to vary without selective forgetting, CLIP tends to overforget. These findings are shown in Tab.\ref{table:aggregated_ablations_appendix} for \textit{AllParamsVary} ablation type and granular results in Tab. \ref{table:vary_allparams}. Therefore, greater control over the varying parameters is necessary in CLIP to achieve forgetting. This differs from the approach used with vision models in \cite{foster_2024_zeroshot}, where all parameters were allowed to vary. This discrepancy can be attributed to the size of the CLIP model, which contains a vast amount of information unlike the smaller models trained on limited data in \cite{foster_2024_zeroshot}, as well as the structural differences that necessitate more controlled updates.

\subsection{Forgetting Using One Iteration}
In \cite{foster_2024_zeroshot} authors used a single epoch for forgetting. In Tab. \ref{table:aggregated_ablations_appendix} in \textit{OneIter} ablation type we can clearly see that this is often not enough to forget as the model most of the time still maintains a very high accuracy on the class to forget. Thus, multiple iterations are required to achieve a desirable level of forgetting. Tab. \ref{table:forget_oneepoch} contains more granular results.

\subsection{Perturbation to Text Embedding}
We analyze how adding noise to the text embeddings, which are in continuous space, rather than using image input changes the forgetting results. Comparing the results in Tab. \ref{table:ablation_textperturb} for \textit{TextEmbPeturb} ablation type and \textit{Original} we see that perturbing textual embedding makes forgetting slightly worse, which is likely due to the fact that perturbing textual embeddings is less meaningful compared to image pixels or discrete text tokens.

\clearpage

\begin{table}[!h]
\onecolumn  \caption{Different ablations. \textit{Original}: results with our method (Lip) from Tab.\ref{table:forget_synth_aggreg}. \textit{AllParamsVary}: forgetting on synthetic data allowing all parameters to vary. \textit{OneIter}: forgetting on synthetic data with a single epoch. \textit{NoTextLoss}: ULip forgetting on synthetic data. \textit{CuPLGen}: synthetic samples generated with CuPL templates.  \textit{EvalTemplChange}: evaluation with different templates. }
    \centering
    \fontsize{20}{25}\selectfont
    \setlength{\tabcolsep}{4pt} 
    \resizebox{1.\textwidth}{!}{\begin{tabular}{@{}lc|cc|cc|cc|cc|cc|cc|cc|c@{}} 
    \toprule 
    \multirow{2}{*}{\makecell[c]{\textbf{Ablation} \\ \textbf{Type}}} & \multirow{2}{*}{\makecell[c]{\textbf{Method}}} & \multicolumn{2}{c}{\makecell[c]{\textbf{Avg. Target} \\ \textbf{Class acc.}}} & \multicolumn{2}{c}{\makecell[c]{\textbf{Avg. Other} \\ \textbf{Classes acc.}}} & \multicolumn{2}{c}{\makecell[c]{\textbf{Avg.} \\ \textbf{StanfordCars}}} & \multicolumn{2}{c}{\makecell[c]{\textbf{Avg.} \\ \textbf{StanfordDogs}}} & \multicolumn{2}{c}{\makecell[c]{\textbf{Avg.} \\ \textbf{Caltech101}}} & \multicolumn{2}{c}{\makecell[c]{\textbf{Avg.} \\ \textbf{OxfordFlowers}}} & \multirow{1}{*}{\makecell[c]{\textbf{Avg. Score} ($\downarrow$)}} \\
    \cline{3-15}
    & & \textbf{BF} & \textbf{AF} & \textbf{BF} & \textbf{AF} & \textbf{BF} & \textbf{AF} & \textbf{BF} & \textbf{AF} & \textbf{BF} & \textbf{AF} & \textbf{BF} & \textbf{AF} \\
    \midrule
    Original & Lip & 0.669 & 0.046 & 0.648 & 0.644 & 0.558 & 0.557 & 0.517 & 0.514 &  0.857 & 0.865  & 0.661 & 0.655 & \textbf{0.018} \\
    \midrule
    AllParamsVary & Lip & 0.669 & 0.0 & 0.648 & 0.033 & 0.558 & 0.015 & 0.517 & 0.014 &  0.857 & 0.091  & 0.661 & 0.011 & 0.954 \\
    \midrule
    OneIter & Lip & 0.655 & 0.173 & 0.629 & 0.631 & 0.558 & 0.561 & 0.517 & 0.522 &  0.857 & 0.869  & 0.661 & 0.662 & 0.053 \\
    \midrule
    NoTextLoss & ULip & 0.669 & 0.606 & 0.648 & 0.641 & 0.558 & 0.557 & 0.517 & 0.511 &  0.857 & 0.854  & 0.661 & 0.651 & 0.189 \\ 
    \midrule
    CuPLGen & Lip & 0.669 & 0.038 & 0.648 & 0.609 & 0.558 & 0.535 & 0.517 & 0.494 &  0.857 & 0.838  & 0.661 & 0.625 & 0.055 \\
    \midrule
    EvalTemplChange & Lip & 0.522 & 0.133 & 0.544 & 0.538 & 0.492 & 0.494 & 0.412 & 0.414 &  0.81 & 0.801  & 0.518 & 0.519 & 0.068 \\
    
    \bottomrule
    \end{tabular}}
   
    \label{table:aggregated_ablations_appendix}
\end{table}

\begin{table}[!h]
\onecolumn \caption{Ablations on perturbation to text embedding. }
    \centering
    \fontsize{20}{25}\selectfont
    \setlength{\tabcolsep}{4pt} 
    \resizebox{1.\textwidth}{!}{\begin{tabular}{@{}cl|cc|cc|cc|cc|cc|cc|cc|c@{}} 
    \toprule 
    \multirow{2}{*}{\makecell[c]{\textbf{Ablation Type}}} & \multirow{2}{*}{\textbf{Dataset}} & \multicolumn{2}{c}{\makecell[c]{\textbf{Avg. Target} \\ \textbf{Class acc.}}} & \multicolumn{2}{c}{\makecell[c]{\textbf{Avg. Other} \\ \textbf{Classes acc.}}} & \multicolumn{2}{c}{\makecell[c]{\textbf{Avg.} \\ \textbf{StanfordCars}}} & \multicolumn{2}{c}{\makecell[c]{\textbf{Avg.} \\ \textbf{StanfordDogs}}} & \multicolumn{2}{c}{\makecell[c]{\textbf{Avg.} \\ \textbf{Caltech101}}} & \multicolumn{2}{c}{\makecell[c]{\textbf{Avg.} \\ \textbf{OxfordFlowers}}} & \multirow{1}{*}{\makecell[c]{\textbf{Avg. Score} ($\downarrow$)}} \\
    \cline{3-15}
    & & \textbf{BF} & \textbf{AF} & \textbf{BF} & \textbf{AF} & \textbf{BF} & \textbf{AF} & \textbf{BF} & \textbf{AF} & \textbf{BF} & \textbf{AF} & \textbf{BF} & \textbf{AF} \\
    \midrule
  
    Original & StanfordCars & 0.397 & 0.056 & 0.558 & 0.551 & - & - & 0.517 & 0.513 &  0.857 & 0.86  & 0.661 & 0.653 & \textbf{0.034} \\
    \midrule
    Original & StanfordDogs & 0.593 & 0.048 & 0.516 & 0.516 & 0.558 & 0.558 & - & - &  0.857 & 0.866  & 0.661 & 0.655 & \textbf{0.018} \\
    \midrule
    Original & Caltech101 & 0.839 & 0.081 & 0.857 & 0.865 & 0.558 & 0.557 & 0.517 & 0.52 &  - & -  & 0.661 & 0.657 & \textbf{0.021} \\
    \midrule
    Original & OxfordFlowers & 0.848 & 0.0 & 0.659 & 0.645 & 0.558 & 0.557 & 0.517 & 0.51 &  0.857 & 0.868  & - & - & \textbf{0.008} \\
    \midrule
    \midrule
    TextEmbPeturb & StanfordCars & 0.397 & 0.056 & 0.558 & 0.55 & - & - & 0.517 & 0.513 &  0.857 & 0.859  & 0.661 & 0.652 & 0.035 \\
    \midrule
    TextEmbPeturb & StanfordDogs & 0.593 & 0.267 & 0.516 & 0.509 & 0.558 & 0.557 & - & - &  0.857 & 0.864  & 0.661 & 0.648 & 0.097 \\
    \midrule
    TextEmbPeturb & Caltech101 & 0.839 & 0.114 & 0.857 & 0.866 & 0.558 & 0.558 & 0.517 & 0.515 &  - & -  & 0.661 & 0.652 & 0.031 \\
    \midrule
    TextEmbPeturb & OxfordFlowers & 0.848 & 0.02 & 0.659 & 0.631 & 0.558 & 0.553 & 0.517 & 0.503 &  0.857 & 0.867  & - & - & 0.02 \\
    \bottomrule
    \end{tabular}}
   
    \label{table:ablation_textperturb}
\end{table}

\begin{table}[!h]
 \caption{We perform forgetting on synthetic samples generated with randomly selected CuPL templates. }
    \centering
    \fontsize{20}{25}\selectfont
    \setlength{\tabcolsep}{4pt} 
    \resizebox{1.\textwidth}{!}{\begin{tabular}{@{}cl|cc|cc|cc|cc|cc|cc|cc|c@{}} 
    \toprule 
    \multirow{2}{*}{\makecell[c]{\textbf{Method}}} & \multirow{2}{*}{\textbf{Dataset}} & \multicolumn{2}{c}{\makecell[c]{\textbf{Avg. Target} \\ \textbf{Class acc.}}} & \multicolumn{2}{c}{\makecell[c]{\textbf{Avg. Other} \\ \textbf{Classes acc.}}} &  \multicolumn{2}{c}{\makecell[c]{\textbf{Avg.} \\ \textbf{StanfordCars}}} & \multicolumn{2}{c}{\makecell[c]{\textbf{Avg.} \\ \textbf{StanfordDogs}}} & \multicolumn{2}{c}{\makecell[c]{\textbf{Avg.} \\ \textbf{Caltech101}}} & \multicolumn{2}{c}{\makecell[c]{\textbf{Avg.} \\ \textbf{OxfordFlowers}}} & \multirow{1}{*}{\makecell[c]{\textbf{Avg. Score} ($\downarrow$)}} \\
    \cline{3-15}
    & & \textbf{BF} & \textbf{AF} & \textbf{BF} & \textbf{AF} & \textbf{BF} & \textbf{AF} & \textbf{BF} & \textbf{AF} & \textbf{BF} & \textbf{AF} & \textbf{BF} & \textbf{AF} \\
    \midrule
    Lip & StanfordCars & 0.397 & 0.032 & 0.558 & 0.525 & - & - & 0.517 & 0.494 &  0.857 & 0.844  & 0.661 & 0.635 & 0.047 \\
    \midrule
    Lip & StanfordDogs & 0.593 & 0.066 & 0.516 & 0.478 & 0.558 & 0.532 & - & - &  0.857 & 0.838  & 0.661 & 0.604 & 0.068  \\
    \midrule
    Lip & Caltech101 & 0.839 & 0.014 & 0.857 & 0.859 & 0.558 & 0.548 & 0.517 & 0.508 &  - & -  & 0.661 & 0.637 & 0.017 \\
    \midrule
    Lip & OxfordFlowers & 0.848 & 0.039 & 0.659 & 0.575 & 0.558 & 0.525 & 0.517 & 0.48 &  0.857 & 0.831  & - & - & 0.067 \\
    \bottomrule
    \end{tabular}}
   
    \label{table:forget_synth_cupl}
\end{table}

\begin{table}[!h]
 \caption{We aggregate across three different evaluation templates to assess sensitivity of models after forgetting to the change in evaluation template. }
    \centering
    \fontsize{20}{25}\selectfont
    \setlength{\tabcolsep}{4pt} 
    \resizebox{1.\textwidth}{!}{\begin{tabular}{@{}cl|cc|cc|cc|cc|cc|cc|cc|c@{}} 
    \toprule 
    \multirow{2}{*}{\makecell[c]{\textbf{Method}}} & \multirow{2}{*}{\textbf{Dataset}} & \multicolumn{2}{c}{\makecell[c]{\textbf{Avg. Target} \\ \textbf{Class acc.}}} & \multicolumn{2}{c}{\makecell[c]{\textbf{Avg. Other} \\ \textbf{Classes acc.}}} &  \multicolumn{2}{c}{\makecell[c]{\textbf{Avg.} \\ \textbf{StanfordCars}}} & \multicolumn{2}{c}{\makecell[c]{\textbf{Avg.} \\ \textbf{StanfordDogs}}} & \multicolumn{2}{c}{\makecell[c]{\textbf{Avg.} \\ \textbf{Caltech101}}} & \multicolumn{2}{c}{\makecell[c]{\textbf{Avg.} \\ \textbf{OxfordFlowers}}} & \multirow{1}{*}{\makecell[c]{\textbf{Avg. Score} ($\downarrow$)}} \\
    \cline{3-15}
    & & \textbf{BF} & \textbf{AF} & \textbf{BF} & \textbf{AF} & \textbf{BF} & \textbf{AF} & \textbf{BF} & \textbf{AF} & \textbf{BF} & \textbf{AF} & \textbf{BF} & \textbf{AF} \\
    \toprule
    Lip & StanfordCars & 0.272 & 0.048 & 0.493 & 0.494 & - & - & 0.412 & 0.413 &  0.81 & 0.802  & 0.518 & 0.512 & 0.041 \\
    \midrule
    Lip & StanfordDogs & 0.306 & 0.103 & 0.416 & 0.397 & 0.492 & 0.491 & - & - &  0.81 & 0.795  & 0.518 & 0.504 & 0.087 \\
    \midrule
    Lip & Caltech101 & 0.897 & 0.211 & 0.809 & 0.809 & 0.492 & 0.499 & 0.412 & 0.422 &  - & -  & 0.518 & 0.547 & 0.047 \\
    \midrule
    Lip & OxfordFlowers & 0.698 & 0.187 & 0.518 & 0.51 & 0.492 & 0.493 & 0.412 & 0.41 &  0.81 & 0.805  & - & - & 0.06 \\    
    \bottomrule
    \end{tabular}}
   
    \label{table:forget_synth_changetemplate}
\end{table}

\begin{table}[!h]
 \caption{Forgetting with Lipschitz loss and synthetic data using a single epoch.}
    \centering
    \fontsize{20}{25}\selectfont
    \setlength{\tabcolsep}{4pt} 
    \resizebox{1.\textwidth}{!}{\begin{tabular}{@{}cl|cc|cc|cc|cc|cc|cc|cc|c@{}} 
    \toprule 
    \multirow{2}{*}{\makecell[c]{\textbf{Method}}} & \multirow{2}{*}{\textbf{Dataset}} & \multicolumn{2}{c}{\makecell[c]{\textbf{Avg. Target} \\ \textbf{Class acc.}}} & \multicolumn{2}{c}{\makecell[c]{\textbf{Avg. Other} \\ \textbf{Classes acc.}}} &  \multicolumn{2}{c}{\makecell[c]{\textbf{Avg.} \\ \textbf{StanfordCars}}} & \multicolumn{2}{c}{\makecell[c]{\textbf{Avg.} \\ \textbf{StanfordDogs}}} & \multicolumn{2}{c}{\makecell[c]{\textbf{Avg.} \\ \textbf{Caltech101}}} & \multicolumn{2}{c}{\makecell[c]{\textbf{Avg.} \\ \textbf{OxfordFlowers}}} & \multirow{1}{*}{\makecell[c]{\textbf{Avg. Score} ($\downarrow$)}} \\
    \cline{3-16}
    & & \textbf{BF} & \textbf{AF} & \textbf{BF} & \textbf{AF} & \textbf{BF} & \textbf{AF} & \textbf{BF} & \textbf{AF} & \textbf{BF} & \textbf{AF} & \textbf{BF} & \textbf{AF} \\

    \midrule
    Lip & StanfordCars & 0.397 & 0.008 & 0.558 & 0.556 & - & - & 0.517 & 0.524 &  0.857 & 0.87  & 0.661 & 0.665 & 0.005 \\
    \midrule
    Lip & StanfordDogs & 0.593 & 0.201 & 0.516 & 0.527 & 0.558 & 0.56 & - & - &  0.857 & 0.87  & 0.661 & 0.66 & 0.068 \\
    \midrule
    Lip & Caltech101 & 0.845 & 0.405 & 0.857 & 0.866 & 0.558 & 0.56 & 0.517 & 0.521 &  - & -  & 0.661 & 0.662 & 0.084 \\
    \midrule
    Lip & OxfordFlowers & 0.848 & 0.189 & 0.659 & 0.653 & 0.558 & 0.562 & 0.517 & 0.519 &  0.857 & 0.867  & - & - & 0.046 \\

    \bottomrule
    \end{tabular}}
   
    \label{table:forget_oneepoch}
\end{table}

\begin{table}[!h]
 \caption{Forgetting with Lipschitz loss and synthetic data when varying all the parameters.}
    \centering
    \fontsize{20}{25}\selectfont
    \setlength{\tabcolsep}{4pt} 
    \resizebox{1.\textwidth}{!}{\begin{tabular}{@{}cl|cc|cc|cc|cc|cc|cc|cc|c@{}} 
    \toprule 
    \multirow{2}{*}{\makecell[c]{\textbf{Method}}} & \multirow{2}{*}{\textbf{Dataset}} & \multicolumn{2}{c}{\makecell[c]{\textbf{Avg. Target} \\ \textbf{Class acc.}}} & \multicolumn{2}{c}{\makecell[c]{\textbf{Avg. Other} \\ \textbf{Classes acc.}}} &  \multicolumn{2}{c}{\makecell[c]{\textbf{Avg.} \\ \textbf{StanfordCars}}} & \multicolumn{2}{c}{\makecell[c]{\textbf{Avg.} \\ \textbf{StanfordDogs}}} & \multicolumn{2}{c}{\makecell[c]{\textbf{Avg.} \\ \textbf{Caltech101}}} & \multicolumn{2}{c}{\makecell[c]{\textbf{Avg.} \\ \textbf{OxfordFlowers}}} & \multirow{1}{*}{\makecell[c]{\textbf{Avg. Score} ($\downarrow$)}}  \\
    \cline{3-16}
    & & \textbf{BF} & \textbf{AF} & \textbf{BF} & \textbf{AF} & \textbf{BF} & \textbf{AF} & \textbf{BF} & \textbf{AF} & \textbf{BF} & \textbf{AF} & \textbf{BF} & \textbf{AF} \\

    \midrule
    Lip & StanfordCars & 0.397 & 0.0 & 0.558 & 0.032 & - & - & 0.517 & 0.018 &  0.857 & 0.114  & 0.661 & 0.008 & 0.753 \\
    \midrule
    Lip & StanfordDogs & 0.593 & 0.0 & 0.516 & 0.009 & 0.558 & 0.013 & - & - &  0.857 & 0.074  & 0.661 & 0.011 & 0.771 \\
    \midrule
    Lip & Caltech101 & 0.839 & 0.0 & 0.857 & 0.073 & 0.558 & 0.014 & 0.517 & 0.012 &  - & -  & 0.661 & 0.013 & 0.769 \\
    \midrule
    Lip & OxfordFlowers & 0.848 & 0.0 & 0.659 & 0.019 & 0.558 & 0.017 & 0.517 & 0.013 &  0.857 & 0.086  & - & - & 0.763 \\

    \bottomrule
    \end{tabular}}
   
    \label{table:vary_allparams}
\end{table}

\begin{table}[!h]
 \caption{Forgetting with ULip loss and synthetic data.}
    \centering
    \fontsize{20}{25}\selectfont
    \setlength{\tabcolsep}{4pt} 
    \resizebox{1.\textwidth}{!}{\begin{tabular}{@{}cl|cc|cc|cc|cc|cc|cc|cc|c@{}} 
    \toprule 
    \multirow{2}{*}{\makecell[c]{\textbf{Method}}} & \multirow{2}{*}{\textbf{Dataset}} & \multicolumn{2}{c}{\makecell[c]{\textbf{Avg. Target} \\ \textbf{Class acc.}}} & \multicolumn{2}{c}{\makecell[c]{\textbf{Avg. Other} \\ \textbf{Classes acc.}}} &  \multicolumn{2}{c}{\makecell[c]{\textbf{Avg.} \\ \textbf{StanfordCars}}} & \multicolumn{2}{c}{\makecell[c]{\textbf{Avg.} \\ \textbf{StanfordDogs}}} & \multicolumn{2}{c}{\makecell[c]{\textbf{Avg.} \\ \textbf{Caltech101}}} & \multicolumn{2}{c}{\makecell[c]{\textbf{Avg.} \\ \textbf{OxfordFlowers}}} & \multirow{1}{*}{\makecell[c]{\textbf{Avg. Score} ($\downarrow$)}}  \\
    \cline{3-15}
    & & \textbf{BF} & \textbf{AF} & \textbf{BF} & \textbf{AF} & \textbf{BF} & \textbf{AF} & \textbf{BF} & \textbf{AF} & \textbf{BF} & \textbf{AF} & \textbf{BF} & \textbf{AF} \\

    \midrule
    ULip & StanfordCars & 0.397 & 0.222 & 0.558 & 0.545 & - & - & 0.517 & 0.504 &  0.857 & 0.849  & 0.661 & 0.648  & 0.128 \\
    \midrule
    ULip & StanfordDogs  & 0.593 & 0.534 & 0.516 & 0.512 & 0.558 & 0.558 & - & - &  0.857 & 0.856  & 0.661 & 0.651 & 0.185 \\
    \midrule
    ULip & Caltech101  & 0.839 & 0.859 & 0.857 & 0.855 & 0.558 & 0.557 & 0.517 & 0.515 &  - & -  & 0.661 & 0.655 & 0.208 \\
    \midrule
    ULip & OxfordFlowers & 0.848 & 0.809 & 0.659 & 0.654 & 0.558 & 0.556 & 0.517 & 0.513 &  0.857 & 0.857  & - & - & 0.194 \\

    \bottomrule
    \end{tabular}}
   
    \label{table:ulip_synth}
\end{table}

\clearpage
\twocolumn

\section{Lip Unlearning Real vs Synthetic}
\label{app:synth_real_rn50}

\begin{table}[!h]
 \onecolumn\caption{Forgetting results with \textbf{synthetic} data using Lipschitz loss. We show the forgetting results on three classes for four different datasets. }
    \centering
    \fontsize{18}{25}\selectfont
    \setlength{\tabcolsep}{4pt} 
    \resizebox{1.\textwidth}{!}{\begin{tabular}{@{}lc|cc|cc|cc|cc|cc|cc|cc|c@{}} 
    \toprule 
    \multirow{2}{*}{\textbf{Dataset}} & \multirow{2}{*}{\textbf{Class name}} & \multicolumn{2}{c}{\makecell[c]{\textbf{Target} \\ \textbf{Class acc.}}} & \multicolumn{2}{c}{\makecell[c]{\textbf{Other} \\ \textbf{Classes acc.}}} & \multicolumn{2}{c}{\makecell[c]{\textbf{Target} \\ \textbf{Class acc.}}} & \multicolumn{2}{c}{\textbf{StanfordCars}} & \multicolumn{2}{c}{\textbf{StanfordDogs}} & \multicolumn{2}{c}{\textbf{Caltech101}} & \multicolumn{2}{c}{\textbf{OxfordFlowers}} & \multirow{1}{*}{\makecell[c]{\textbf{Avg. Score} ($\downarrow$)}} \\
    \cline{3-17}
    & & \textbf{BF} & \textbf{AF} & \textbf{BF} & \textbf{AF} & {\makecell[c]{\textbf{Synt.} \\ \textbf{train}}} & {\makecell[c]{\textbf{Real} \\ \textbf{valid.}}} & \textbf{BF} & \textbf{AF} & \textbf{BF} & \textbf{AF} & \textbf{BF} & \textbf{AF} & \textbf{BF} & \textbf{AF}\\
    \midrule
    StanfordDogs & Pekinese & 0.705 & 0.066 & 0.515 & 0.514 & 0.062 & 0.0  & 0.558 & 0.559 & - & - &  0.857 & 0.867  & 0.661 & 0.658 & 0.02 \\
    StanfordDogs & toy poodle & 0.574 & 0.033 & 0.516 & 0.518 & 0.031 & 0.0  & 0.558 & 0.559 & - & - &  0.857 & 0.867  & 0.661 & 0.647 & 0.016 \\
    StanfordDogs & Scotch terrier & 0.5 & 0.047 & 0.517 & 0.516 & 0.047 & 0.083  & 0.558 & 0.557 & - & - &  0.857 & 0.865  & 0.661 & 0.66 & 0.019 \\
    StanfordCars & 2009 Spyker C8 Coupe & 0.262 & 0.024 & 0.559 & 0.553 & 0.0 & 0.0  & - & - & 0.517 & 0.518 &  0.857 & 0.865  & 0.661 & 0.66 & 0.021 \\
    StanfordCars & 2010 Dodge Ram Pickup 3500 Crew Cab & 0.405 & 0.143 & 0.558 & 0.544 & 0.0 & 0.0  & - & - & 0.517 & 0.502 &  0.857 & 0.845  & 0.661 & 0.638 & 0.091 \\
    StanfordCars & 2011 Ford Ranger SuperCab & 0.524 & 0.0 & 0.558 & 0.555 & 0.0 & 0.0  & - & - & 0.517 & 0.52 &  0.857 & 0.869  & 0.661 & 0.661  & 0.001 \\
    Caltech101 & euphonium & 0.789 & 0.0 & 0.858 & 0.868 & 0.016 & 0.0  & 0.558 & 0.557 & 0.517 & 0.52 &  - & -  & 0.661 & 0.658 & 0.001 \\
    Caltech101 & minaret & 0.826 & 0.043 & 0.857 & 0.863 & 0.0 & 0.067  & 0.558 & 0.556 & 0.517 & 0.515 &  - & -  & 0.661 & 0.661 & 0.012 \\
    Caltech101 & platypus & 0.9 & 0.2 & 0.857 & 0.866 & 0.062 & 0.286  & 0.558 & 0.558 & 0.517 & 0.524 &  - & -  & 0.661 & 0.653 & 0.047 \\
    OxfordFlowers & gazania & 0.957 & 0.0 & 0.658 & 0.649 & 0.062 & 0.0  & 0.558 & 0.559 & 0.517 & 0.513 &  0.857 & 0.869  & - & - & 0.004 \\
    OxfordFlowers & tree mallow & 1.0 & 0.0 & 0.658 & 0.643 & 0.047 & 0.0  & 0.558 & 0.557 & 0.517 & 0.51 &  0.857 & 0.869  & - & - & 0.011 \\
    OxfordFlowers & trumpet creeper & 0.588 & 0.0 & 0.661 & 0.643 & 0.047 & 0.083  & 0.558 & 0.557 & 0.517 & 0.503 &  0.857 & 0.866  & - & -  & 0.011 \\
    \bottomrule
    \end{tabular}}
   
    \label{table:forget_synth}
\end{table}

\begin{table}[!h]
\caption{Forgetting results with \textbf{real} data using Lipschitz loss. We show the forgetting results on three classes for four different datasets.}
    \centering
    \fontsize{18}{25}\selectfont
    \setlength{\tabcolsep}{4pt} 
    \resizebox{1.\textwidth}{!}{\begin{tabular}{@{}lc|cc|cc|c|cc|cc|cc|cc|c@{}} 
    \toprule 
    \multirow{2}{*}{\textbf{Dataset}} & \multirow{2}{*}{\textbf{Class name}} & \multicolumn{2}{c}{\makecell[c]{\textbf{Target} \\ \textbf{Class acc.}}} & \multicolumn{2}{c}{\makecell[c]{\textbf{Other} \\ \textbf{Classes acc.}}} & \multicolumn{1}{c}{\makecell[c]{\textbf{Target} \\ \textbf{Class acc.}}} & \multicolumn{2}{c}{\textbf{StanfordCars}} & \multicolumn{2}{c}{\textbf{StanfordDogs}} & \multicolumn{2}{c}{\textbf{Caltech101}} & \multicolumn{2}{c}{\textbf{OxfordFlowers}} & \multirow{1}{*}{\makecell[c]{\textbf{Avg. Score} ($\downarrow$)}} \\
    \cline{3-16}
    & & \textbf{BF} & \textbf{AF} & \textbf{BF} & \textbf{AF} & {\makecell[c]{\textbf{Real} \\ \textbf{valid.}}} & \textbf{BF} & \textbf{AF} & \textbf{BF} & \textbf{AF} & \textbf{BF} & \textbf{AF} & \textbf{BF} & \textbf{AF} \\
    \midrule 
    StanfordDogs & Pekinese & 0.705 & 0.066 & 0.515 & 0.514 &  0.045 & 0.558 & 0.563 & - & - &  0.857 & 0.873  & 0.661 & 0.654 & 0.021 \\
    StanfordDogs & toy poodle & 0.574 & 0.033 & 0.516 & 0.506 &  0.022 & 0.558 & 0.565 & - & - &  0.857 & 0.871  & 0.661 & 0.646 & 0.02 \\
    StanfordDogs & Scotch terrier & 0.5 & 0.016 & 0.517 & 0.509 &  0.043 & 0.558 & 0.562 & - & - &  0.857 & 0.862  & 0.661 & 0.654 & 0.011 \\
    StanfordCars & 2009 Spyker C8 Coupe & 0.262 & 0.024 & 0.559 & 0.517 &  0.059 & - & - & 0.517 & 0.509 &  0.857 & 0.849  & 0.661 & 0.642 & 0.044  \\
    StanfordCars & 2010 Dodge Ram Pickup 3500 Crew Cab & 0.405 & 0.024 & 0.558 & 0.557 &  0.0 & - & - & 0.517 & 0.52 &  0.857 & 0.866  & 0.661 & 0.655 & 0.014 \\
    StanfordCars & 2011 Ford Ranger SuperCab & 0.524 & 0.048 & 0.558 & 0.561 &  0.029 & - & - & 0.517 & 0.521 &  0.857 & 0.87  & 0.661 & 0.658 & 0.019 \\
    Caltech101 & euphonium & 0.789 & 0.0 & 0.858 & 0.853 &  0.0 & 0.558 & 0.549 & 0.517 & 0.495 &  - & -  & 0.661 & 0.625 & 0.023 \\
    Caltech101 & minaret & 0.826 & 0.043 & 0.857 & 0.865 &  0.026 & 0.558 & 0.563 & 0.517 & 0.51 &  - & -  & 0.661 & 0.657 & 0.014 \\
    Caltech101 & platypus & 0.9 & 0.5 & 0.857 & 0.866 &  0.235 & 0.558 & 0.559 & 0.517 & 0.517 &  - & -  & 0.661 & 0.652 & 0.114 \\
    OxfordFlowers & gazania & 0.957 & 0.0 & 0.658 & 0.65 &  0.026 & 0.558 & 0.561 & 0.517 & 0.515 &  0.857 & 0.866  & - & - & 0.003 \\
    OxfordFlowers & tree mallow & 1.0 & 0.353 & 0.658 & 0.652 &  0.0 & 0.558 & 0.56 & 0.517 & 0.512 &  0.857 & 0.861  & - & - & 0.074 \\
    OxfordFlowers & trumpet creeper & 0.588 & 0.059 & 0.661 & 0.658 &  0.069 & 0.558 & 0.56 & 0.517 & 0.515 &  0.857 & 0.864  & - & - & 0.022 \\
    \bottomrule
    \end{tabular}}
    \label{table:forget_real}
\end{table}

\clearpage
\twocolumn

\section{Forgetting on Multiple Classes and Error Analysis}
\label{app:err_mcls}

\label{sec:forget_multiplecls}
We assess forgetting on multiple classes in Tab. \ref{table:forget_synth_multiclass_rn50_appendix}. \textit{Other Classes acc.} represents the performance on the remaining classes before forgetting (BF) and after forgetting (AF) respectively on the dataset specified in \textit{Dataset}. We observe that after forgetting multiple classes, the accuracy on remaining classes and other datasets remains relatively high on StanfordDog and Caltech101 while reducing slightly more on other two datasets. This is due to the 1.5\% loss in accuracy on other classes when unlearning \textit{2010 Dodge Ram Pickup 3500 Crew Cab} and 2\% when unlearning \textit{trumpet creeper} alone as shown in Tab. \ref{table:full_forget_synth} for LIP unlearning, leading to more forgetting subsequently on other classes when forgetting multiple classes. This does not happen on StanfordDogs and Caltech101 datasets where the reduction in performance after forgetting on not targeted classes is low. Indeed, replacing \textit{2010 Dodge Ram Pickup 3500 Crew Cab} with \textit{2012 Rolls-Royce Ghost Sedan} that has a smaller reduction in performance on other classes after forgetting as shown in Tab.\ref{table:forget_other}, forgetting on multiple classes also improves. These are shown on the last row in Tab.\ref{table:forget_other}. In general, it is normal to expect that as the number of classes to forget increases, the accuracy on other classes will gradually decrease, as demonstrated in our experiments, albeit at a relatively slow rate. 

Notably, on the \textit{OxfordFlowers} dataset, the accuracy decrease is more pronounced compared to other datasets when forgetting was applied, indicating higher sensitivity. This observation held true for single class forgetting as well. Therefore, we examined how close in terms of their position in the sorted list of logits scores the correct prediction of the original model and incorrect prediction of the forget model were before and after forgetting. If their proximity in terms of position and logits scores was significant, it would suggest that the model was already uncertain about those predictions. Consequently, a minor, targeted forgetting may have resulted in a subtle change in logits scores, swapping the predictions.

We conducted this analysis on the model after forgetting the selected classes. Looking at Tab. \ref{table:forget_synth_multiclass_rn50_appendix}, following the forgetting of three classes from \textit{Caltech101} datasets, there were 106 incorrect predictions after forgetting compared to the model's performance before forgetting on the \textit{OxfordFlowers} dataset. Among these, we found that in 88 instances (83.02\%), the correct class in an ordered set of logits predicted by CLIP shifted by one place. This means that instead of the correct class having the highest score, it ranked second-highest after forgetting. At the same time, to be included in those 88 instances the new highest incorrect prediction of the forget model must have been the second highest class in the original model's logits.

To explain this procedure with an example, consider a given image of a \textit{Poodle} for which the original model predicted the classes in the following way:
\textit{Poodle, Labrador, Spaniel} ordered by logits assumed to be \textit{15, 14.9, 12}. Here, \textit{Poodle} was the correct prediction with the highest score, as we only examined cases where the original model was correct and assessed how that changed after forgetting. After the forgetting process, the new model predicts: \textit{Labrador, Poodle, Spaniel} with sorted scores \textit{15.1, 14.9, 11.8}. In this case, the correct prediction (\textit{Poodle}) moved from the highest to second highest score, i.e., one step away from its original position, and the difference in scores is 0.02 (corresponding to \textit{One-Step Avg. LogitScore} in Tab. \ref{table:sensit_anal}). Also, the new incorrect prediction with the highest score, \textit{Labrador}, was the second highest prediction in the original model logits, thus this example would be included in calculating the \textit{One-Step \%}. 

Similarly, looking at shifts of up to two places, where the correct class ranked either second or third highest, we found that 99 cases (93.4\%) moved away by a maximum of two places from the correct prediction.

For \textit{StanfordDogs}, there were 120 incorrect labels. In 100 cases (83.33\%), the correct class moved away by one place, and in 112 cases (93.33\%), it shifted by a maximum of two places.

For \textit{StanfordCars}, we took the model after forgetting on classes shown on the last row of Tab. 6. There were 124 incorrect labels. In 104 cases (83.87\%), the correct class moved away by one place, and in 116 cases (93.54\%), it shifted by a maximum of two places. 

These results are summarized in Table \ref{table:sensit_anal}. The $\Delta$ \textit{One-Step Avg. LogitScore} represents the standardized average change in logits scores between the model's new incorrect and correct predictions. Similarly, for $\Delta$ \textit{Two-Step Avg. LogitScore}, considering a shift of two places.

This analysis indicates that the greater than expected drop in accuracy on the \textit{OxfordFlowers} dataset is attributed to the model's original uncertainty about those cases. Thus, despite our demonstrated ability to precisely target the parameters for forgetting the target class, the model's initial uncertainty contributes to the relatively more pronounced decrease in accuracy.

\newpage
\onecolumn\begin{table}[!h]
\caption{Sensitivity Analysis on OxfordFlowers dataset.}
    \centering
    \fontsize{10}{15}\selectfont
    \setlength{\tabcolsep}{4pt} 
    \resizebox{0.8\textwidth}{!}{\begin{tabular}{@{}ccccc@{}} 
    \toprule 
    \textbf{Model} & \textbf{One-Step \%} & $\Delta$ \textbf{One-Step Avg. LogitScore} & \textbf{Two-Step \%} & $\Delta$ \textbf{Two-Step Avg. LogitScore} \\
    StanfordDogs & 83.02 & 0.014 & 93.4 & 0.014 \\
    StanfordCars & 83.87 & 0.015 & 93.54 & 0.016 \\ 
    Caltech101 & 83.33 & 0.015 & 93.33 & 0.017 \\
    
    \bottomrule
    \end{tabular}}
    
    \label{table:sensit_anal}
\end{table}

\begin{table}[!h]
\caption{Forgetting other examples using Lipschitz loss.}
    \centering
    \fontsize{18}{30}\selectfont
    \setlength{\tabcolsep}{4pt} 
    \resizebox{1.\textwidth}{!}{\begin{tabular}{@{}clc|cc|cc|cc|cc|cc|cc|cc|c@{}} 
    \toprule 
    \multirow{2}{*}{\makecell[c]{\textbf{Forgetting} \\ \textbf{Data}}} & \multirow{2}{*}{\textbf{Dataset}} & \multirow{2}{*}{\textbf{Class name}} & \multicolumn{2}{c}{\makecell[c]{\textbf{Target} \\ \textbf{Class acc.}}} & \multicolumn{2}{c}{\makecell[c]{\textbf{Other} \\ \textbf{Classes acc.}}} & \multicolumn{2}{c}{\makecell[c]{\textbf{Target} \\ \textbf{Class acc.}}} & \multicolumn{2}{c}{\textbf{StanfordCars}} & \multicolumn{2}{c}{\textbf{StanfordDogs}} & \multicolumn{2}{c}{\textbf{Caltech101}} & \multicolumn{2}{c}{\textbf{OxfordFlowers}} & \multirow{1}{*}{\makecell[c]{\textbf{Avg. Score  ($\downarrow$)}}} \\
    \cline{4-18}
    & & & \textbf{BF} & \textbf{AF} & \textbf{BF} & \textbf{AF} & {\makecell[c]{\textbf{Synt.} \\ \textbf{train}}} & {\makecell[c]{\textbf{Real} \\ \textbf{valid.}}} & \textbf{BF} & \textbf{AF} & \textbf{BF} & \textbf{AF} & \textbf{BF} & \textbf{AF} & \textbf{BF} & \textbf{AF}\\
    \midrule

    Generated & StanfordCars & 2012 Chevrolet Avalanche Crew Cab & 0.622 & 0.044 & 0.557 & 0.494 & 0.281 & 0.0  & - & - & 0.517 & 0.501 &  0.857 & 0.87  & 0.661 & 0.633  & 0.05 \\
    Generated & StanfordCars & 2012 Rolls-Royce Ghost Sedan & 0.526 & 0.07 & 0.557 & 0.557 & 0.078 & 0.15  & - & - & 0.520 & 0.527 & 0.857 & 0.862 & 0.66 & 0.658 & 0.04 \\
    \midrule
    Generated & StanfordCars & \shortstack{2009 Spyker C8 Coupe,\\2012 Rolls-Royce Ghost Sedan,\\2011 Ford Ranger SuperCab} & 0.397 & 0.12 & 0.558 & 0.535 & - & - & - & - & 0.517 & 0.501 & 0.857 & 0.862 & 0.661 & 0.631 & 0.08 \\

    \bottomrule
    \end{tabular}}
    \label{table:forget_other}
\end{table}

\begin{table}[!h]
\caption{Forgetting multiple classes with generated data using Lipschitz loss.}
    \centering
    \fontsize{18}{30}\selectfont
    \setlength{\tabcolsep}{4pt} 
    \resizebox{1.\textwidth}{!}{\begin{tabular}{@{}lc|cc|cc|cc|cc|cc|cc|c@{}} 
    \toprule 
    \multirow{2}{*}{\textbf{Dataset}} & \multirow{2}{*}{\textbf{Classes}}  & \multicolumn{2}{c}{\makecell[c]{\textbf{Avg. Target} \\ \textbf{Classes acc.}}} & \multicolumn{2}{c}{\makecell[c]{\textbf{Other} \\ \textbf{Classes acc.}}} & \multicolumn{2}{c}{\textbf{StanfordCars}} & \multicolumn{2}{c}{\textbf{StanfordDogs}} & \multicolumn{2}{c}{\textbf{Caltech101}} & \multicolumn{2}{c}{\textbf{OxfordFlowers}} & \multirow{1}{*}{\makecell[c]{\textbf{Avg. Score  ($\downarrow$)}}} \\
    \cline{3-15}
    & & \textbf{BF} & \textbf{AF} & \textbf{BF} & \textbf{AF} & \textbf{BF} & \textbf{AF} & \textbf{BF} & \textbf{AF} & \textbf{BF} & \textbf{AF} & \textbf{BF} & \textbf{AF} \\
    \toprule
    StanfordDogs & Pekinese,toy poodle,Scotch terrier & 0.593 & 0.09 & 0.517 & 0.507 & 0.558 & 0.547 & - & - &  0.857 & 0.865  & 0.661 & 0.633 & 0.046 \\
    \midrule
    StanfordCars & \shortstack{2009 Spyker C8 Coupe,\\2010 Dodge Ram Pickup 3500 Crew Cab,\\2011 Ford Ranger SuperCab} & 0.397 & 0.2 & 0.558 & 0.519 & - & - & 0.517 & 0.482 &  0.857 & 0.84  & 0.661 & 0.607 & 0.16 \\
    \midrule
    Caltech101 & euphonium,minaret,platypus & 0.839 & 0.125 & 0.857 & 0.869 & 0.558 & 0.549 & 0.517 & 0.515 &  - & -  & 0.661 & 0.633 & 0.042 \\
    \midrule
    OxfordFlowers & gazania,tree mallow,trumpet creeper & 0.848 & 0.0 & 0.661 & 0.609 & 0.558 & 0.552 & 0.517 & 0.498 &  0.857 & 0.863  & - & - & 0.023 \\
    \bottomrule
    \end{tabular}}
    \label{table:forget_synth_multiclass_rn50_appendix}
\end{table}

\twocolumn
\clearpage
\section{Synthetic Images Visualization}
\label{app:synth_vis}
In Fig. \ref{fig:synt_images} we can see an example of a synthetic image for a class from four different datasets we tested on. These synthetic samples do not have a clear appearance of the sample class but are enough for unlearning the class. 

\begin{figure}[!b]
\centering
\onecolumn\includegraphics[width=0.9\textwidth]{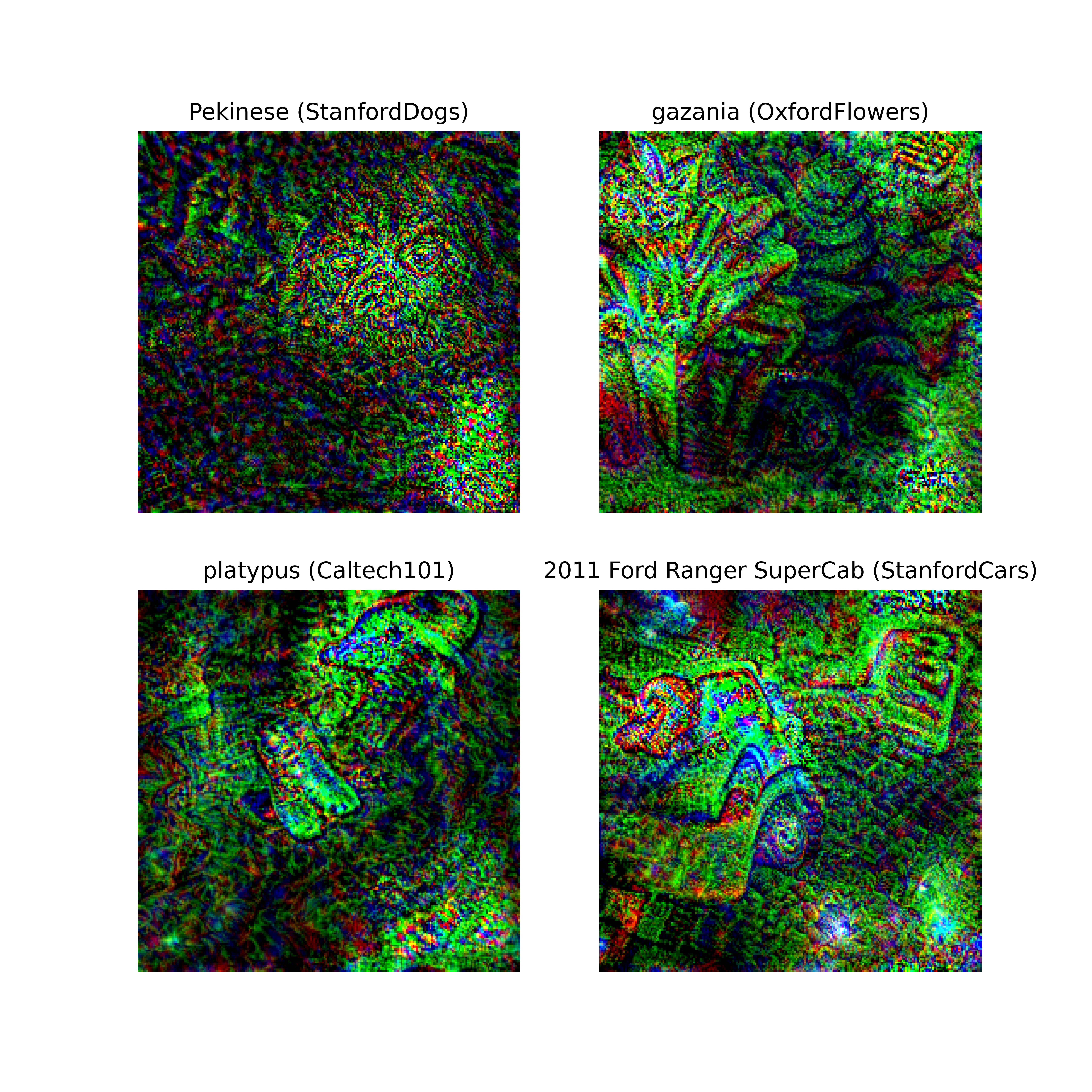}
\caption{Synthetic images examples. }
\label{fig:synt_images}
\end{figure} 

\twocolumn
\clearpage

\twocolumn

\clearpage

\section{ViT Results}
\label{app:vit_results}

Results for CLIP with ViT-B/16 visual encoder are included in Tab. \ref{table:forget_synth_vit}. In general we observe that forgetting with ViT compared to ResNet architecture is harder and the class accuracy after forgetting, even if decreases, it often does not go to zero. This difference may be attributed to the fact that ViT learns more fine-grained representations of the input images compared to ResNet thanks to its self-attention mechanism. As a result, specific classes could be encoded in a more intricate manner within ViT's learned representations, making them harder to unlearn without affecting other aspects of the model's knowledge. When it comes to other methods for comparison the conclusions are similar to the ones highlighted for ResNet. Aggregated results are shown in Tab. \ref{table:forget_synth_aggreg_ViT}

\begin{table}[!b]
 \onecolumn\caption{Aggregated forgetting results with ViT-B/16 visual encoder. We compare our method (Lip) to five other methods averaging across three classes for four selected datasets. We aim to minimize the \textit{Avg. Target Class acc. AF} while maintaining \textit{Avg. Other Classes acc. AF} and other datasets at a similar level to that before forgetting (BF). We bold the best results comparing only among the first four methods that are zero-shot methods for a fair comparison. }
    \centering
    \fontsize{20}{25}\selectfont
    \setlength{\tabcolsep}{4pt} 
    \resizebox{\textwidth}{!}{\begin{tabular}{@{}clc|cc|cc|cc|cc|cc|cc|c@{}} 
    \toprule 
    \multirow{2}{*}{\makecell[c]{\textbf{Method}}} & \multirow{2}{*}{\textbf{Dataset}} & \multirow{2}{*}{\makecell[c]{\textbf{Forgetting} \\ \textbf{Type}}} & \multicolumn{2}{c}{\makecell[c]{\textbf{Avg. Target} \\ \textbf{Class acc.}}} & \multicolumn{2}{c}{\makecell[c]{\textbf{Avg. Other} \\ \textbf{Classes acc.}}} &  \multicolumn{2}{c}{\makecell[c]{\textbf{Avg.} \\ \textbf{StanfordCars}}} & \multicolumn{2}{c}{\makecell[c]{\textbf{Avg.} \\ \textbf{StanfordDogs}}} & \multicolumn{2}{c}{\makecell[c]{\textbf{Avg.} \\ \textbf{Caltech101}}} & \multicolumn{2}{c}{\makecell[c]{\textbf{Avg.} \\ \textbf{OxfordFlowers}}} & \multirow{1}{*}{\makecell[c]{\textbf{Avg. Score} ($\downarrow$)}} \\\\
    \cline{4-16}
    & & & \textbf{BF} & \textbf{AF} & \textbf{BF} & \textbf{AF} & \textbf{BF} & \textbf{AF} & \textbf{BF} & \textbf{AF} & \textbf{BF} & \textbf{AF} & \textbf{BF} & \textbf{AF} \\
    \midrule
    Lip & StanfordCars & ZS & 0.595 & 0.159 & 0.656 & 0.642 & - & - & 0.591 & 0.584 &  0.933 & 0.932  & 0.708 & 0.707 & \textbf{0.06} \\
    Emb & StanfordCars & ZS & 0.595 & 0.0 & 0.656 & 0.557 & - & - & 0.591 & 0.508 &  0.933 & 0.921  & 0.708 & 0.69 & 0.066 \\
    Amns & StanfordCars & ZS & 0.595 & 0.143 & 0.656 & 0.18 & - & - & 0.591 & 0.398 &  0.933 & 0.876  & 0.708 & 0.51 & 0.327 \\
    EMMN & StanfordCars & ZS & 0.595 & 0.159 & 0.656 & 0.182 & - & - & 0.591 & 0.119 &  0.933 & 0.589  & 0.708 & 0.137 & 0.592 \\
    \hdashline
    ULip & StanfordCars & semi ZS & 0.595 & 0.032 & 0.656 & 0.419 & - & - & 0.591 & 0.59 &  0.933 & 0.93  & 0.708 & 0.711 & 0.084 \\
    AmnsRetain & StanfordCars & not ZS & 0.595 & 0.0 & 0.656 & 0.744 & - & - & 0.591 & 0.476 &  0.933 & 0.887  & 0.708 & 0.554 & 0.092 \\
    Salun & StanfordCars & not ZS & 0.595 & 0.087 & 0.656 & 0.711 & - & - & 0.591 & 0.495 &  0.933 & 0.863  & 0.708 & 0.58 & 0.113 \\
    \midrule
    Lip & StanfordDogs & ZS & 0.673 & 0.142 & 0.591 & 0.592 & 0.655 & 0.647 & - & - &  0.933 & 0.935  & 0.708 & 0.709 & \textbf{0.045} \\
    Emb & StanfordDogs & ZS & 0.673 & 0.071 & 0.591 & 0.518 & 0.655 & 0.632 & - & - &  0.933 & 0.93  & 0.708 & 0.699 & 0.056 \\
    Amns & StanfordDogs & ZS & 0.673 & 0.219 & 0.591 & 0.358 & 0.655 & 0.59 & - & - &  0.933 & 0.901  & 0.708 & 0.572 & 0.209 \\
    EMMN & StanfordDogs & ZS & 0.673 & 0.042 & 0.591 & 0.365 & 0.655 & 0.284 & - & - &  0.933 & 0.826  & 0.708 & 0.438 & 0.301 \\
    \hdashline
    ULip & StanfordDogs & semi ZS & 0.673 & 0.588 & 0.591 & 0.536 & 0.655 & 0.649 & - & - &  0.933 & 0.93  & 0.708 & 0.702 & 0.197 \\
    AmnsRetain & StanfordDogs & not ZS & 0.673 & 0.021 & 0.591 & 0.698 & 0.655 & 0.472 & - & - &  0.933 & 0.841  & 0.708 & 0.495 & 0.142 \\
    Salun & StanfordDogs & not ZS & 0.673 & 0.043 & 0.591 & 0.662 & 0.655 & 0.508 & - & - &  0.933 & 0.835  & 0.708 & 0.609 & 0.107 \\
    \midrule
    Lip & Caltech101 & ZS & 0.971 & 0.576 & 0.933 & 0.935 & 0.655 & 0.652 & 0.591 & 0.594 &  - & -  & 0.708 & 0.709 & \textbf{0.12} \\
    Emb & Caltech101 & ZS & 0.971 & 0.598 & 0.933 & 0.91 & 0.655 & 0.609 & 0.591 & 0.517 &  - & -  & 0.708 & 0.656 & 0.182 \\
    Amns & Caltech101 & ZS & 0.971 & 0.846 & 0.933 & 0.848 & 0.655 & 0.517 & 0.591 & 0.445 &  - & -  & 0.708 & 0.533 & 0.334 \\
    EMMN & Caltech101 & ZS & 0.971 & 0.284 & 0.933 & 0.813 & 0.655 & 0.352 & 0.591 & 0.302 &  - & -  & 0.708 & 0.473 & 0.341 \\
    \hdashline
    ULip & Caltech101 & semi ZS & 0.971 & 0.812 & 0.933 & 0.916 & 0.655 & 0.65 & 0.591 & 0.584 &  - & -  & 0.708 & 0.702 & 0.177 \\
    AmnsRetain & Caltech101 & not ZS & 0.971 & 0.0 & 0.933 & 0.946 & 0.655 & 0.506 & 0.591 & 0.474 &  - & -  & 0.708 & 0.544 & 0.132 \\
    Salun & Caltech101 & not ZS & 0.971 & 0.0 & 0.933 & 0.926 & 0.655 & 0.525 & 0.591 & 0.506 &  - & -  & 0.708 & 0.642 & 0.088 \\
    \midrule
    Lip & OxfordFlowers & ZS & 0.784 & 0.078 & 0.707 & 0.702 & 0.655 & 0.645 & 0.591 & 0.588 &  0.933 & 0.933  & - & - & \textbf{0.026} \\
    Emb & OxfordFlowers & ZS & 0.784 & 0.0 & 0.707 & 0.617 & 0.655 & 0.543 & 0.591 & 0.522 &  0.933 & 0.906  & - & - & 0.089 \\
    Amns & OxfordFlowers & ZS & 0.784 & 0.834 & 0.707 & 0.527 & 0.655 & 0.602 & 0.591 & 0.526 &  0.933 & 0.913  & - & - & 0.307 \\
    EMMN & OxfordFlowers & ZS & 0.784 & 0.02 & 0.707 & 0.433 & 0.655 & 0.317 & 0.591 & 0.304 &  0.933 & 0.83  & - & - & 0.305 \\
    \hdashline
    ULip & OxfordFlowers & semi ZS & 0.784 & 0.02 & 0.707 & 0.529 & 0.655 & 0.64 & 0.591 & 0.554 &  0.933 & 0.913  & - & - & 0.077 \\
    AmnsRetain & OxfordFlowers & not ZS & 0.784 & 0.0 & 0.707 & 0.945 & 0.655 & 0.56 & 0.591 & 0.531 &  0.933 & 0.914  & - & - & 0.054 \\
    Salun & OxfordFlowers & not ZS & 0.784 & 0.059 & 0.707 & 0.923 & 0.655 & 0.537 & 0.591 & 0.503 &  0.933 & 0.857  & - & - & 0.097 \\
    \bottomrule
    \end{tabular}}
   
    \label{table:forget_synth_aggreg_ViT}
\end{table}

\begin{table}[!b]
\caption{Forgetting with real data using Lipschitz loss ViT-B/16 visual encoder.}
    \centering
    \fontsize{18}{30}\selectfont
    \setlength{\tabcolsep}{4pt} 
    \resizebox{\textwidth}{!}{\begin{tabular}{@{}lc|cc|cc|c|cc|cc|cc|cc|c@{}} 
    \toprule 
    \multirow{2}{*}{\textbf{Dataset}} & \multirow{2}{*}{\textbf{Class name}} & \multicolumn{2}{c}{\makecell[c]{\textbf{Target} \\ \textbf{Class acc.}}} & \multicolumn{2}{c}{\makecell[c]{\textbf{Other} \\ \textbf{Classes acc.}}} & \multicolumn{1}{c}{\makecell[c]{\textbf{Target} \\ \textbf{Class acc.}}} & \multicolumn{2}{c}{\textbf{StanfordCars}} & \multicolumn{2}{c}{\textbf{StanfordDogs}} & \multicolumn{2}{c}{\textbf{Caltech101}} & \multicolumn{2}{c}{\textbf{OxfordFlowers}} & \multirow{1}{*}{\makecell[c]{\textbf{Avg. Score} ($\downarrow$)}} \\
    \cline{3-16}
    & & \textbf{BF} & \textbf{AF} & \textbf{BF} & \textbf{AF} & {\makecell[c]{\textbf{Real} \\ \textbf{valid.}}} & \textbf{BF} & \textbf{AF} & \textbf{BF} & \textbf{AF} & \textbf{BF} & \textbf{AF} & \textbf{BF} & \textbf{AF} \\
    \midrule 
    StanfordDogs & Pekinese & 0.787 & 0.016 & 0.59 & 0.607 &  0.0 & 0.655 & 0.657 & - & - &  0.933 & 0.932  & 0.708 & 0.709 & 0.004 \\
    StanfordDogs & toy poodle & 0.607 & 0.131 & 0.591 & 0.581 &  0.022 & 0.655 & 0.64 & - & - &  0.933 & 0.94  & 0.708 & 0.715 & 0.051 \\
    StanfordDogs & Scotch terrier & 0.625 & 0.047 & 0.591 & 0.56 &  0.085 & 0.655 & 0.64 & - & - &  0.933 & 0.937  & 0.708 & 0.709 & 0.03 \\
    StanfordCars & 2009 Spyker C8 Coupe & 0.429 & 0.238 & 0.656 & 0.652 &  0.176 & - & - & 0.591 & 0.591 &  0.933 & 0.933  & 0.708 & 0.708 & 0.112 \\
    StanfordCars & 2010 Dodge Ram Pickup 3500 Crew Cab & 0.548 & 0.143 & 0.656 & 0.637 &  0.088 & - & - & 0.591 & 0.598 &  0.933 & 0.938  & 0.708 & 0.713 & 0.058 \\
    StanfordCars & 2011 Ford Ranger SuperCab & 0.81 & 0.81 & 0.654 & 0.654 &  0.794 & - & - & 0.591 & 0.593 &  0.933 & 0.933  & 0.708 & 0.71 & 0.224 \\
    Caltech101 & euphonium & 1.0 & 0.0 & 0.933 & 0.935 &  0.031 & 0.655 & 0.656 & 0.591 & 0.596 &  - & -  & 0.708 & 0.707 & 0.0 \\
    Caltech101 & minaret & 0.913 & 0.739 & 0.933 & 0.928 &  0.711 & 0.655 & 0.644 & 0.591 & 0.593 &  - & -  & 0.708 & 0.719 & 0.167 \\
    Caltech101 & platypus & 1.0 & 0.9 & 0.933 & 0.934 &  0.647 & 0.655 & 0.656 & 0.591 & 0.599 &  - & -  & 0.708 & 0.714  & 0.18 \\
    OxfordFlowers & gazania & 1.0 & 0.087 & 0.705 & 0.711 &  0.077 & 0.655 & 0.653 & 0.591 & 0.601 &  0.933 & 0.934  & - & - & 0.018 \\
    OxfordFlowers & tree mallow & 0.765 & 0.294 & 0.707 & 0.7 &  0.172 & 0.655 & 0.652 & 0.591 & 0.601 &  0.933 & 0.941  & - & - & 0.08 \\
    OxfordFlowers & trumpet creeper & 0.588 & 0.118 & 0.709 & 0.713 &  0.241 & 0.655 & 0.648 & 0.591 & 0.602 &  0.933 & 0.939  & - & - & 0.042 \\
    \bottomrule
    \end{tabular}}
    
    \label{table:forget_real_vit}
\end{table}

\begin{table}[!b]
\caption{Forgetting on multiple classes using Lipschitz loss ViT-B/16 visual encoder.}
    \centering
    \fontsize{18}{30}\selectfont
    \setlength{\tabcolsep}{4pt} 
    \resizebox{1.\textwidth}{!}{\begin{tabular}{@{}lc|cc|cc|cc|cc|cc|cc|c@{}} 
    \toprule 
    \multirow{2}{*}{\textbf{Dataset}} & \multirow{2}{*}{\textbf{Classes}}  & \multicolumn{2}{c}{\makecell[c]{\textbf{Avg. Target} \\ \textbf{Classes acc.}}} & \multicolumn{2}{c}{\makecell[c]{\textbf{Other} \\ \textbf{Classes acc.}}} & \multicolumn{2}{c}{\textbf{StanfordCars}} & \multicolumn{2}{c}{\textbf{StanfordDogs}} & \multicolumn{2}{c}{\textbf{Caltech101}} & \multicolumn{2}{c}{\textbf{OxfordFlowers}} & \multirow{1}{*}{\makecell[c]{\textbf{Avg. Score} ($\downarrow$)}}  \\
    \cline{3-15}
    & & \textbf{BF} & \textbf{AF} & \textbf{BF} & \textbf{AF} & \textbf{BF} & \textbf{AF} & \textbf{BF} & \textbf{AF} & \textbf{BF} & \textbf{AF} & \textbf{BF} & \textbf{AF} \\
    \midrule
    StanfordDogs & Pekinese,toy poodle,Scotch terrier & 0.672 & 0.251 & 0.589 & 0.584 & 0.655 & 0.644 & - & - &  0.933 & 0.939  & 0.708 & 0.713 & 0.08 \\
    StanfordCars & \shortstack{2009 Spyker C8 Coupe,\\2010 Dodge Ram Pickup 3500 Crew Cab,\\2011 Ford Ranger SuperCab} & 0.595 & 0.3 & 0.656 & 0.625 & - & - & 0.591 & 0.576 &  0.933 & 0.928  & 0.708 & 0.699 & 0.119 \\
    Caltech101 & euphonium,minaret,platypus & 0.971 & 0.498 & 0.932 & 0.929 & 0.655 & 0.634 & 0.591 & 0.589 &  - & -  & 0.708 & 0.709 & 0.11 \\
    OxfordFlowers & trumpet creeper,gazania,tree mallow & 0.807 & 0.31 & 0.705 & 0.68 & 0.655 & 0.613 & 0.591 & 0.551 &  0.933 & 0.929  & - & - & 0.111 \\
    \bottomrule
    \end{tabular}}
    \label{table:forget_synth_multiclass_vit}
\end{table}

\clearpage
\begin{table}[!b]
\onecolumn
\caption{Forgetting results with ViT-B/16 visual encoder. We compare our methods with five others on three classes for four selected datasets. We bold the best results comparing only among the first four methods that are zero-shot methods for a fair comparison. }
    \centering
    \fontsize{25}{25}\selectfont
    \setlength{\tabcolsep}{4pt} 
    \resizebox{1.\textwidth}{!}{\begin{tabular}{@{}clc|cc|cc|cc|cc|cc|cc|
    cc|c@{}} 
    \toprule 
    \multirow{2}{*}{\makecell[c]{\textbf{Method}}} & \multirow{2}{*}{\textbf{Dataset}} & \multirow{2}{*}{\textbf{Class name}} & \multicolumn{2}{c}{\makecell[c]{\textbf{Target} \\ \textbf{Class acc.}}} & \multicolumn{2}{c}{\makecell[c]{\textbf{Other} \\ \textbf{Classes acc.}}} & \multicolumn{2}{c}{\makecell[c]{\textbf{Target} \\ \textbf{Class acc.}}} & \multicolumn{2}{c}{\textbf{StanfordCars}} & \multicolumn{2}{c}{\textbf{StanfordDogs}} & \multicolumn{2}{c}{\textbf{Caltech101}} & \multicolumn{2}{c}{\textbf{OxfordFlowers}} & \multirow{1}{*}{\makecell[c]{\textbf{Avg. Score  ($\downarrow$)}}}  \\
    \cline{4-18}
    & & & \textbf{BF} & \textbf{AF} & \textbf{BF} & \textbf{AF} & {\makecell[c]{\textbf{Synt.} \\ \textbf{train}}} & {\makecell[c]{\textbf{Real} \\ \textbf{valid.}}} & \textbf{BF} & \textbf{AF} & \textbf{BF} & \textbf{AF} & \textbf{BF} & \textbf{AF} & \textbf{BF} & \textbf{AF}\\
    \midrule

    Lip & StanfordDogs & Pekinese & 0.787 & 0.377 & 0.59 & 0.601 & 0.094 & 0.273  & 0.655 & 0.656 & - & - &  0.933 & 0.934  & 0.708 & 0.708 & 0.096 \\
    Lip & StanfordDogs & toy poodle & 0.607 & 0.033 & 0.591 & 0.593 & 0.0 & 0.0  & 0.655 & 0.639 & - & - &  0.933 & 0.932  & 0.708 & 0.707 & \textbf{0.016} \\
    Lip & StanfordDogs & Scotch terrier & 0.625 & 0.016 & 0.591 & 0.582 & 0.219 & 0.0  & 0.655 & 0.647 & - & - &  0.933 & 0.938  & 0.708 & 0.713 & \textbf{0.011} \\
    Lip & StanfordCars & 2009 Spyker C8 Coupe & 0.429 & 0.262 & 0.656 & 0.639 & 0.484 & 0.222  & - & - & 0.591 & 0.581 &  0.933 & 0.93  & 0.708 & 0.7 & \textbf{0.134} \\
    Lip & StanfordCars & 2010 Dodge Ram Pickup 3500 Crew Cab & 0.548 & 0.048 & 0.656 & 0.634 & 0.062 & 0.0  & - & - & 0.591 & 0.58 &  0.933 & 0.933  & 0.708 & 0.708 & 0.028 \\
    Lip & StanfordCars & 2011 Ford Ranger SuperCab & 0.81 & 0.167 & 0.654 & 0.653 & 0.984 & 0.125  & - & - & 0.591 & 0.59 &  0.933 & 0.933  & 0.708 & 0.713 & \textbf{0.042} \\
    Lip & Caltech101 & euphonium & 1.0 & 0.158 & 0.933 & 0.935 & 0.0 & 0.154  & 0.655 & 0.653 & 0.591 & 0.597 &  - & -  & 0.708 & 0.706 & \textbf{0.033} \\
    Lip & Caltech101 & minaret & 0.913 & 0.87 & 0.933 & 0.932 & 0.031 & 1.0  & 0.655 & 0.649 & 0.591 & 0.59 &  - & -  & 0.708 & 0.709 & \textbf{0.187} \\
    Lip & Caltech101 & platypus & 1.0 & 0.7 & 0.933 & 0.936 & 0.031 & 0.857  & 0.655 & 0.653 & 0.591 & 0.595 &  - & -  & 0.708 & 0.711 & \textbf{0.141} \\
    Lip & OxfordFlowers & gazania & 1.0 & 0.0 & 0.705 & 0.7 & 0.297 & 0.0  & 0.655 & 0.642 & 0.591 & 0.587 &  0.933 & 0.935  & - & - & 0.007 \\
    Lip & OxfordFlowers & tree mallow & 0.765 & 0.176 & 0.707 & 0.699 & 0.172 & 0.0  & 0.655 & 0.65 & 0.591 & 0.596 &  0.933 & 0.933  & - & - & \textbf{0.05} \\
    Lip & OxfordFlowers & trumpet creeper & 0.588 & 0.059 & 0.709 & 0.705 & 0.781 & 0.0  & 0.655 & 0.644 & 0.591 & 0.581 &  0.933 & 0.932  & - & - & \textbf{0.028} \\
    
    \midrule

    Emb & StanfordDogs & Pekinese & 0.787 & 0.213 & 0.59 & 0.601 & 0.031 & 0.227  & 0.655 & 0.656 & - & - &  0.933 & 0.934  & 0.708 & 0.708 & \textbf{0.054} \\
    Emb & StanfordDogs & toy poodle & 0.607 & 0.0 & 0.591 & 0.472 & 0.0 & 0.0  & 0.655 & 0.621 & - & - &  0.933 & 0.931  & 0.708 & 0.696 & 0.054 \\
    Emb & StanfordDogs & Scotch terrier & 0.625 & 0.0 & 0.591 & 0.481 & 0.141 & 0.0  & 0.655 & 0.617 & - & - &  0.933 & 0.926  & 0.708 & 0.695 & 0.054 \\
    Emb & StanfordCars & 2009 Spyker C8 Coupe & 0.429 & 0.0 & 0.656 & 0.479 & 0.016 & 0.0  & - & - & 0.591 & 0.392 &  0.933 & 0.908  & 0.708 & 0.659 & 0.14 \\
    Emb & StanfordCars & 2010 Dodge Ram Pickup 3500 Crew Cab & 0.548 & 0.0 & 0.656 & 0.626 & 0.078 & 0.0  & - & - & 0.591 & 0.59 &  0.933 & 0.934  & 0.708 & 0.713 & \textbf{0.01} \\
    Emb & StanfordCars & 2011 Ford Ranger SuperCab & 0.81 & 0.0 & 0.654 & 0.565 & 0.203 & 0.0  & - & - & 0.591 & 0.542 &  0.933 & 0.92  & 0.708 & 0.699 & 0.049 \\
    Emb & Caltech101 & euphonium & 1.0 & 0.368 & 0.933 & 0.935 & 0.0 & 0.385  & 0.655 & 0.652 & 0.591 & 0.594 &  - & -  & 0.708 & 0.709 & 0.075 \\
    Emb & Caltech101 & minaret & 0.913 & 0.826 & 0.933 & 0.933 & 0.016 & 0.867  & 0.655 & 0.635 & 0.591 & 0.583 &  - & -  & 0.708 & 0.711 & 0.19 \\
    Emb & Caltech101 & platypus & 1.0 & 0.6 & 0.933 & 0.861 & 0.0 & 0.429  & 0.655 & 0.539 & 0.591 & 0.376 &  - & -  & 0.708 & 0.547 & 0.289 \\
    Emb & OxfordFlowers & gazania & 1.0 & 0.0 & 0.705 & 0.705 & 0.141 & 0.0  & 0.655 & 0.645 & 0.591 & 0.593 &  0.933 & 0.933  & - & - & \textbf{0.003} \\
    Emb & OxfordFlowers & tree mallow & 0.765 & 0.0 & 0.707 & 0.577 & 0.172 & 0.0  & 0.655 & 0.58 & 0.591 & 0.501 &  0.933 & 0.903  & - & - & 0.097 \\
    Emb & OxfordFlowers & trumpet creeper & 0.588 & 0.0 & 0.709 & 0.569 & 0.0 & 0.0  & 0.655 & 0.406 & 0.591 & 0.472 &  0.933 & 0.88  & - & - & 0.167 \\
    \midrule

    Amns & StanfordDogs & Pekinese & 0.787 & 0.623 & 0.59 & 0.366 & 0.031 & 0.545  & 0.655 & 0.581 & - & - &  0.933 & 0.896  & 0.708 & 0.609 & 0.293 \\
    Amns & StanfordDogs & toy poodle & 0.607 & 0.033 & 0.591 & 0.234 & 0.0 & 0.0  & 0.655 & 0.57 & - & - &  0.933 & 0.899  & 0.708 & 0.482 & 0.229 \\
    Amns & StanfordDogs & Scotch terrier & 0.625 & 0.0 & 0.591 & 0.473 & 0.062 & 0.042  & 0.655 & 0.618 & - & - &  0.933 & 0.908  & 0.708 & 0.626 & 0.08 \\
    Amns & StanfordCars & 2009 Spyker C8 Coupe & 0.429 & 0.0 & 0.656 & 0.058 & 0.0 & 0.0  & - & - & 0.591 & 0.242 &  0.933 & 0.808  & 0.708 & 0.361 & 0.425 \\
    Amns & StanfordCars & 2010 Dodge Ram Pickup 3500 Crew Cab & 0.548 & 0.214 & 0.656 & 0.166 & 0.0 & 0.0  & - & - & 0.591 & 0.436 &  0.933 & 0.904  & 0.708 & 0.572 & 0.325 \\
    Amns & StanfordCars & 2011 Ford Ranger SuperCab & 0.81 & 0.214 & 0.654 & 0.315 & 0.094 & 0.25  & - & - & 0.591 & 0.516 &  0.933 & 0.916  & 0.708 & 0.596 & 0.217 \\
    Amns & Caltech101 & euphonium & 1.0 & 1.0 & 0.933 & 0.901 & 0.406 & 0.923  & 0.655 & 0.648 & 0.591 & 0.57 &  - & -  & 0.708 & 0.639 & 0.236 \\
    Amns & Caltech101 & minaret & 0.913 & 0.739 & 0.933 & 0.774 & 0.094 & 0.6  & 0.655 & 0.336 & 0.591 & 0.257 &  - & -  & 0.708 & 0.366 & 0.503 \\
    Amns & Caltech101 & platypus & 1.0 & 0.8 & 0.933 & 0.868 & 0.078 & 0.714  & 0.655 & 0.566 & 0.591 & 0.507 &  - & -  & 0.708 & 0.594 & 0.262 \\
    Amns & OxfordFlowers & gazania & 1.0 & 0.913 & 0.705 & 0.518 & 0.062 & 0.875  & 0.655 & 0.586 & 0.591 & 0.514 &  0.933 & 0.908  & - & - & 0.288 \\
    Amns & OxfordFlowers & tree mallow & 0.765 & 0.824 & 0.707 & 0.484 & 0.094 & 0.75  & 0.655 & 0.593 & 0.591 & 0.513 &  0.933 & 0.91  & - & - & 0.329 \\
    Amns & OxfordFlowers & trumpet creeper & 0.588 & 0.765 & 0.709 & 0.578 & 0.094 & 0.75  & 0.655 & 0.627 & 0.591 & 0.55 &  0.933 & 0.92  & - & - & 0.322 \\

    \midrule
    EMMN & StanfordDogs & Pekinese & 0.787 & 0.0 & 0.59 & 0.376 & - & -  & 0.655 & 0.278 & - & - &  0.933 & 0.828  & 0.708 & 0.432 & 0.288 \\
    EMMN & StanfordDogs & toy poodle & 0.607 & 0.0 & 0.591 & 0.373 & - & -  & 0.655 & 0.308 & - & - &  0.933 & 0.836  & 0.708 & 0.446 & 0.275 \\
    EMMN & StanfordDogs & Scotch terrier & 0.625 & 0.125 & 0.591 & 0.347 & - & -  & 0.655 & 0.265 & - & - &  0.933 & 0.813  & 0.708 & 0.436 & 0.344 \\
    EMMN & StanfordCars & 2009 Spyker C8 Coupe & 0.429 & 0.0 & 0.656 & 0.188 & - & -  & - & - & 0.591 & 0.116 &  0.933 & 0.614  & 0.708 & 0.148 & 0.53 \\
    EMMN & StanfordCars & 2010 Dodge Ram Pickup 3500 Crew Cab & 0.548 & 0.476 & 0.656 & 0.184 & - & -  & - & - & 0.591 & 0.13 &  0.933 & 0.56  & 0.708 & 0.126 & 0.718 \\
    EMMN & StanfordCars & 2011 Ford Ranger SuperCab & 0.81 & 0.0 & 0.654 & 0.175 & - & -  & - & - & 0.591 & 0.111 &  0.933 & 0.594  & 0.708 & 0.136 & 0.543 \\
    EMMN & Caltech101 & euphonium & 1.0 & 0.105 & 0.933 & 0.783 & - & -  & 0.655 & 0.352 & 0.591 & 0.297 &  - & -  & 0.708 & 0.45 & 0.318 \\
    EMMN & Caltech101 & minaret & 0.913 & 0.348 & 0.933 & 0.817 & - & -  & 0.655 & 0.36 & 0.591 & 0.315 &  - & -  & 0.708 & 0.485 & 0.347 \\
    EMMN & Caltech101 & platypus & 1.0 & 0.4 & 0.933 & 0.838 & - & -  & 0.655 & 0.345 & 0.591 & 0.294 &  - & -  & 0.708 & 0.484 & 0.359 \\
    EMMN & OxfordFlowers & gazania & 1.0 & 0.0 & 0.705 & 0.44 & - & -  & 0.655 & 0.308 & 0.591 & 0.312 &  0.933 & 0.832  & - & - & 0.297 \\
    EMMN & OxfordFlowers & tree mallow & 0.765 & 0.0 & 0.707 & 0.445 & - & -  & 0.655 & 0.33 & 0.591 & 0.288 &  0.933 & 0.829  & - & - & 0.298 \\
    EMMN & OxfordFlowers & trumpet creeper & 0.588 & 0.059 & 0.709 & 0.413 & - & -  & 0.655 & 0.312 & 0.591 & 0.31 &  0.933 & 0.828  & - & - & 0.326 \\
    
    \midrule

    AmnsRetain & StanfordDogs & Pekinese & 0.787 & 0.0 & 0.59 & 0.706 & - & - & 0.655 & 0.473 & - & - &  0.933 & 0.829  & 0.708 & 0.49 & 0.139 \\
    AmnsRetain & StanfordDogs & toy poodle & 0.607 & 0.0 & 0.591 & 0.709 & - & - & 0.655 & 0.475 & - & - &  0.933 & 0.847  & 0.708 & 0.507 & 0.13 \\
    AmnsRetain & StanfordDogs & Scotch terrier & 0.625 & 0.062 & 0.591 & 0.679 & - & - & 0.655 & 0.468 & - & - &  0.933 & 0.848  & 0.708 & 0.488 & 0.158 \\
    AmnsRetain & StanfordCars & 2009 Spyker C8 Coupe & 0.429 & 0.0 & 0.656 & 0.768 & - & - & - & - & 0.591 & 0.475 &  0.933 & 0.88  & 0.708 & 0.573 & 0.089 \\
    AmnsRetain & StanfordCars & 2010 Dodge Ram Pickup 3500 Crew Cab & 0.548 & 0.0 & 0.656 & 0.742 & - & - & - & - & 0.591 & 0.494 &  0.933 & 0.89  & 0.708 & 0.549 & 0.087 \\
    AmnsRetain & StanfordCars & 2011 Ford Ranger SuperCab & 0.81 & 0.0 & 0.654 & 0.721 & - & - & - & - & 0.591 & 0.46 &  0.933 & 0.891  & 0.708 & 0.539 & 0.101 \\
    AmnsRetain & Caltech101 & euphonium & 1.0 & 0.0 & 0.933 & 0.945 & - & - & 0.655 & 0.493 & 0.591 & 0.46 &  - & -  & 0.708 & 0.556 & 0.137 \\
    AmnsRetain & Caltech101 & minaret & 0.913 & 0.0 & 0.933 & 0.938 & - & - & 0.655 & 0.494 & 0.591 & 0.468 &  - & -  & 0.708 & 0.532 & 0.14 \\
    AmnsRetain & Caltech101 & platypus & 1.0 & 0.0 & 0.933 & 0.956 & - & - & 0.655 & 0.53 & 0.591 & 0.493 &  - & -  & 0.708 & 0.543 & 0.118 \\
    AmnsRetain & OxfordFlowers & gazania & 1.0 & 0.0 & 0.705 & 0.944 & - & - & 0.655 & 0.563 & 0.591 & 0.522 &  0.933 & 0.916  & - & - & 0.055 \\
    AmnsRetain & OxfordFlowers & tree mallow & 0.765 & 0.0 & 0.707 & 0.934 & - & - & 0.655 & 0.558 & 0.591 & 0.528 &  0.933 & 0.913  & - & - & 0.055 \\
    AmnsRetain & OxfordFlowers & trumpet creeper & 0.588 & 0.0 & 0.709 & 0.956 & - & - & 0.655 & 0.558 & 0.591 & 0.542 &  0.933 & 0.913  & - & - & 0.051 \\
    
    \midrule

    Salun & StanfordDogs & Pekinese & 0.787 & 0.049 & 0.59 & 0.669 & - & - & 0.655 & 0.521 & - & - &  0.933 & 0.842  & 0.708 & 0.605 & 0.102 \\
    Salun & StanfordDogs & toy poodle & 0.607 & 0.066 & 0.591 & 0.654 & - & - & 0.655 & 0.508 & - & - &  0.933 & 0.838  & 0.708 & 0.627 & 0.11 \\
    Salun & StanfordDogs & Scotch terrier & 0.625 & 0.016 & 0.591 & 0.663 & - & - & 0.655 & 0.495 & - & - &  0.933 & 0.824  & 0.708 & 0.594 & 0.109 \\
    Salun & StanfordCars & 2009 Spyker C8 Coupe & 0.429 & 0.167 & 0.656 & 0.718 & - & - & - & - & 0.591 & 0.5 &  0.933 & 0.869  & 0.708 & 0.583 & 0.157 \\
    Salun & StanfordCars & 2010 Dodge Ram Pickup 3500 Crew Cab & 0.548 & 0.048 & 0.656 & 0.709 & - & - & - & - & 0.591 & 0.497 &  0.933 & 0.855  & 0.708 & 0.563 & 0.107 \\
    Salun & StanfordCars & 2011 Ford Ranger SuperCab & 0.81 & 0.048 & 0.654 & 0.707 & - & - & - & - & 0.591 & 0.488 &  0.933 & 0.865  & 0.708 & 0.592 & 0.094 \\
    Salun & Caltech101 & euphonium & 1.0 & 0.0 & 0.933 & 0.928 & - & - & 0.655 & 0.531 & 0.591 & 0.504 &  - & -  & 0.708 & 0.654 & 0.084 \\
    Salun & Caltech101 & minaret & 0.913 & 0.0 & 0.933 & 0.927 & - & - & 0.655 & 0.517 & 0.591 & 0.509 &  - & -  & 0.708 & 0.639 & 0.091 \\
    Salun & Caltech101 & platypus & 1.0 & 0.0 & 0.933 & 0.923 & - & - & 0.655 & 0.528 & 0.591 & 0.506 &  - & -  & 0.708 & 0.633 & 0.091 \\
    Salun & OxfordFlowers & gazania & 1.0 & 0.0 & 0.705 & 0.922 & - & - & 0.655 & 0.525 & 0.591 & 0.506 &  0.933 & 0.86  & - & - & 0.084 \\
    Salun & OxfordFlowers & tree mallow & 0.765 & 0.059 & 0.707 & 0.929 & - & - & 0.655 & 0.542 & 0.591 & 0.498 &  0.933 & 0.858  & - & - & 0.098 \\
    Salun & OxfordFlowers & trumpet creeper & 0.588 & 0.118 & 0.709 & 0.919 & - & - & 0.655 & 0.542 & 0.591 & 0.506 &  0.933 & 0.854  & - & - & 0.12 \\

    \midrule

    ULip & StanfordDogs & Pekinese & 0.787 & 0.836 & 0.59 & 0.569 & - & - & 0.655 & 0.651 & - & - &  0.933 & 0.933  & 0.708 & 0.709 & 0.221 \\
    ULip & StanfordDogs & toy poodle & 0.607 & 0.098 & 0.591 & 0.508 & - & - & 0.655 & 0.646 & - & - &  0.933 & 0.927  & 0.708 & 0.694 & 0.069 \\
    ULip & StanfordDogs & Scotch terrier & 0.625 & 0.828 & 0.591 & 0.531 & - & - & 0.655 & 0.649 & - & - &  0.933 & 0.931  & 0.708 & 0.704 & 0.289 \\
    ULip & StanfordCars & 2009 Spyker C8 Coupe & 0.429 & 0.048 & 0.656 & 0.494 & - & - & - & - & 0.591 & 0.591 &  0.933 & 0.932  & 0.708 & 0.716 & 0.072 \\
    ULip & StanfordCars & 2010 Dodge Ram Pickup 3500 Crew Cab & 0.548 & 0.0 & 0.656 & 0.515 & - & - & - & - & 0.591 & 0.598 &  0.933 & 0.931  & 0.708 & 0.713 & 0.043 \\
    ULip & StanfordCars & 2011 Ford Ranger SuperCab & 0.81 & 0.048 & 0.654 & 0.247 & - & - & - & - & 0.591 & 0.581 &  0.933 & 0.927  & 0.708 & 0.704 & 0.142 \\
    ULip & Caltech101 & euphonium & 1.0 & 1.0 & 0.933 & 0.919 & - & - & 0.655 & 0.645 & 0.591 & 0.59 &  - & -  & 0.708 & 0.705 & 0.207 \\
    ULip & Caltech101 & minaret & 0.913 & 0.435 & 0.933 & 0.923 & - & - & 0.655 & 0.655 & 0.591 & 0.59 &  - & -  & 0.708 & 0.707 & 0.098 \\
    ULip & Caltech101 & platypus & 1.0 & 1.0 & 0.933 & 0.907 & - & - & 0.655 & 0.65 & 0.591 & 0.571 &  - & -  & 0.708 & 0.694 & 0.218 \\
    ULip & OxfordFlowers & gazania & 1.0 & 0.0 & 0.705 & 0.404 & - & - & 0.655 & 0.626 & 0.591 & 0.539 &  0.933 & 0.898  & - & - & 0.12 \\
    ULip & OxfordFlowers & tree mallow & 0.765 & 0.34 & 0.707 & 0.672 & - & - & 0.655 & 0.653 & 0.591 & 0.572 &  0.933 & 0.925  & - & - & 0.06 \\
    ULip & OxfordFlowers & trumpet creeper & 0.588 & 0.059 & 0.709 & 0.51 & - & - & 0.655 & 0.641 & 0.591 & 0.552 &  0.933 & 0.915  & - & - & 0.097 \\
    \bottomrule
    \end{tabular}}

    \label{table:forget_synth_vit}
\end{table}

\clearpage
\twocolumn
\section{Unlearning Identities}
\label{app:face_unlearn}

In line with our motivation of the right to be forgotten, we assess whether our unlearning method is able to forget faces. For this experiment we choose PinsFaces \cite{pinsfaces} dataset that contains 105 celebrity faces. As we can see in Tabs. \ref{table:unlearn_faces_full_rn50} and \ref{table:unlearn_faces_full_vit} our method is also valid to unlearn faces.

\begin{table}[!h]
\onecolumn\caption{Unlearning Faces Full Results with RN50.}
    \centering
    \fontsize{18}{30}\selectfont
    \setlength{\tabcolsep}{4pt} 
    \resizebox{0.8\textwidth}{!}{\begin{tabular}{@{}lc|cc|cc|cc|cc|cc|cc|c@{}} 
    \toprule 
    \multirow{2}{*}{\textbf{Dataset}} & \multirow{2}{*}{\textbf{Class name}} & \multicolumn{2}{c}{\makecell[c]{\textbf{Target} \\ \textbf{Class acc.}}} & \multicolumn{2}{c}{\makecell[c]{\textbf{Other} \\ \textbf{Classes acc.}}} &  \multicolumn{2}{c}{\textbf{StanfordCars}} & \multicolumn{2}{c}{\textbf{StanfordDogs}} & \multicolumn{2}{c}{\textbf{Caltech101}} & \multicolumn{2}{c}{\textbf{OxfordFlowers}} & \multirow{1}{*}{\makecell[c]{\textbf{Avg. Score  ($\downarrow$)}}} \\
    \cline{3-15}
    & & \textbf{BF} & \textbf{AF} & \textbf{BF} & \textbf{AF} & \textbf{BF} & \textbf{AF} & \textbf{BF} & \textbf{AF} & \textbf{BF} & \textbf{AF} & \textbf{BF} & \textbf{AF} \\
    \midrule 
    PinsFaces & Gal Gadot & 0.707 & 0.03 & 0.822 & 0.8 & 0.558 & 0.55 & 0.517 & 0.503 &  0.857 & 0.863  & 0.661 & 0.642 & 0.029  \\
    PinsFaces & Henry Cavil & 0.866 & 0.062 & 0.82 & 0.781 & 0.558 & 0.545 & 0.517 & 0.492 &  0.857 & 0.857  & 0.661 & 0.623 & 0.049 \\
    PinsFaces & Amanda Crew & 0.828 & 0.069 & 0.821 & 0.815  & 0.558 & 0.557 & 0.517 & 0.511 &  0.857 & 0.861  & 0.661 & 0.649 & 0.024 \\
    \bottomrule
    \end{tabular}}
    
    \label{table:unlearn_faces_full_rn50}
\end{table}

\begin{table}[!h]
\onecolumn\caption{Unlearning Faces Full Results with ViT-B/16.}
    \centering
    \fontsize{18}{30}\selectfont
    \setlength{\tabcolsep}{4pt} 
    \resizebox{0.8\textwidth}{!}{\begin{tabular}{@{}lc|cc|cc|cc|cc|cc|cc|c@{}} 
    \toprule 
    \multirow{2}{*}{\textbf{Dataset}} & \multirow{2}{*}{\textbf{Class name}} & \multicolumn{2}{c}{\makecell[c]{\textbf{Target} \\ \textbf{Class acc.}}} & \multicolumn{2}{c}{\makecell[c]{\textbf{Other} \\ \textbf{Classes acc.}}} &  \multicolumn{2}{c}{\textbf{StanfordCars}} & \multicolumn{2}{c}{\textbf{StanfordDogs}} & \multicolumn{2}{c}{\textbf{Caltech101}} & \multicolumn{2}{c}{\textbf{OxfordFlowers}} & \multirow{1}{*}{\makecell[c]{\textbf{Avg. Score  ($\downarrow$)}}} \\
    \cline{3-15}
    & & \textbf{BF} & \textbf{AF} & \textbf{BF} & \textbf{AF} & \textbf{BF} & \textbf{AF} & \textbf{BF} & \textbf{AF} & \textbf{BF} & \textbf{AF} & \textbf{BF} & \textbf{AF} \\
    \midrule 

    PinsFaces & Gal Gadot & 0.879 & 0.091 & 0.908 & 0.908 & 0.655 & 0.65 & 0.591 & 0.589 &  0.933 & 0.932  & 0.708 & 0.708 & 0.023 \\
    PinsFaces & Henry Cavil & 0.959 & 0.216 & 0.907 & 0.903 & 0.655 & 0.646 & 0.591 & 0.585 &  0.933 & 0.932  & 0.708 & 0.695 & 0.055 \\
    PinsFaces & Amanda Crew & 1.0 & 0.0 & 0.907 & 0.91 & 0.655 & 0.656 & 0.591 & 0.594 &  0.933 & 0.933  & 0.708 & 0.71 & 0.0 \\

    \bottomrule
    \end{tabular}}
    
    \label{table:unlearn_faces_full_vit}
\end{table}

\clearpage
\section{Forgetting Algorithm}
\label{app:forg_algo}

\begin{algorithm}
\caption{CLIP forgetting}
\label{alg:mainalgo}
\begin{algorithmic}
\Require CLIP image and text encoders: $f_{\theta}$, $f_{\phi}$
\Require Textual class to forget: $x_{text}$, Learning rate: $\alpha$, 
\Require Initial number of visual and textual layers to update: $InitV_{up}, InitT_{up}$
\Require Increase steps sigma: $s$, visual layers: $v$, textual layers: $t$
\Require Number of perturbations: $N$, Initial Gaussian perturbation: $\sigma$
\Require Accuracy stopping threshold: $GoalAcc$, Total increase steps: $I$
\Require Optimizer: $optim(\theta, \phi, lr=\alpha)$
\LineComment{\parbox[t]{1\linewidth}{Function that generates synthetic training samples by gradient ascent as in Eq. 5}} 
\State $X \gets SyntheticImagesGen(x_{text})$ 
\For{$n$ in $range(I)$}    
    
    \For{$x_{img}$ in $X$}
        \State $\ell = 0$
        \For{$i$ in $range(N)$}
            \State Sample $\epsilon \sim \mathcal{N}(0,\sigma^2)$
            \State $x'_{img} = x_{img} + \epsilon$
            \State $k = \frac{\|f_\theta(\boldsymbol{x_{img}})-f_\theta(\boldsymbol{x'_{img}})\|_2 + \|f_\phi(\boldsymbol{x_{text}})-f_\theta(\boldsymbol{x'_{img}})\|_2}{\|\epsilon\|_2}$
        
        \State $\ell = \ell + k$
        \EndFor

        \State $\ell = \ell / N$
        \LineComment{\parbox[t]{.9\linewidth}{Update $InitV_{up}$ and $InitT_{up}$ most important layers in visual and textual branch respectively}}
        \State $\theta, \phi \gets SelectiveUpdate(optim\{\Delta_{\theta, \phi} \ell\}, InitV_{up}, InitT_{up})$ 
        
        \State $Acc \gets EvalAcc(X)$ \Comment{\parbox[t]{.6\linewidth}{Accuracy on synthetic training samples}}
        \If{$Acc < GoalAcc$}
        \State \Return $\theta, \phi$
        \Else 
        \LineComment{\parbox[t]{.9\linewidth}{Increase parameters to reduce more aggressively the accuracy on synthetic samples}}
        \State $InitV_{up} \gets InitV_{up} + v$\
        \State $InitT_{up} \gets InitT_{up} + t$\
        \State $\sigma \gets \sigma + s$\
        \EndIf
    \EndFor
\EndFor
\State \Return $\theta, \phi$
\end{algorithmic}
\end{algorithm}

\clearpage
\twocolumn
\section{Verification of Forgetting Success and Data Generation Threshold}
\label{app:forg_success}

\paragraph{Data Generation Threshold}
We observe that generating samples classified as the target class alone is not sufficient for effective forgetting. To achieve successful forgetting, the generated samples must exhibit a high probability of belonging to the forget class, otherwise the forgetting process might fail. Empirically, we find that selecting a threshold around 0.7 works well for ResNet50 visual encoder and around 0.9 for ViT-B/16 visual encoder, but sometimes more tuning of the threshold is required. 

\paragraph{Verification of Forgetting Success}
The accuracy achieved on synthetic forget data should serve as a measure of how effectively the model has forgotten a class. We find that for this indicator to be consistent the probability of the predicted class on synthetic samples need to be close to the probability of the real samples, otherwise there might be some discrepancy. For example, in Tab. \ref{table:forget_mismatch} in the 1st row \textit{2009 Bentley Arnage Sedan} class maintains a high accuracy on \textit{Synth. train} data despite \textit{Target Class acc. AF} dropping to 5\%. Conversely, in the 2nd row we observe that for \textit{revolver} class the accuracy on \textit{Synt. train} subset is low, yet forgetting was not successful as indicated by the true validation accuracy \textit{Target Class acc. AF Real valid.}. It turns out that for \textit{revolver} class the probability of the generated samples is too low while for \textit{2009 Bentley Arnage Sedan} is too high. In most cases, we adopted a threshold of 0.7 in the standard setting. Thus, by generating synthetic samples with class predicted probability closer to that of real samples we reduce the discrepancy as demonstrated in the last two rows of Tab.\ref{table:forget_mismatch}. Note that generating higher probability synthetic samples compared to probability of real samples would still generally suffice to forget the class but not to verify the forgetting success. 
A simple alternative is to rely on a small real validation subset of the class to forget or if available, stored past probabilities of the class prediction. 
This verification can be conducted by a user if they are unwilling to share their data with the company. 

\begin{table}[!h]
\onecolumn\caption{Forgetting verification discrepancy examples.}
    \centering
    \fontsize{18}{30}\selectfont
    \setlength{\tabcolsep}{4pt} 
    \resizebox{1.\textwidth}{!}{\begin{tabular}{@{}clc|cc|cc|cc|cc|cc|cc|cc@{}} 
    \toprule 
    \multirow{2}{*}{\makecell[c]{\textbf{Discrepancy}}} & \multirow{2}{*}{\textbf{Dataset}} & \multirow{2}{*}{\textbf{Class name}} & \multicolumn{2}{c}{\makecell[c]{\textbf{Target} \\ \textbf{Class acc.}}} & \multicolumn{2}{c}{\makecell[c]{\textbf{Other} \\ \textbf{Classes acc.}}} & \multicolumn{2}{c}{\makecell[c]{\textbf{Target} \\ \textbf{Class acc.}}} & \multicolumn{2}{c}{\textbf{StanfordCars}} & \multicolumn{2}{c}{\textbf{StanfordDogs}} & \multicolumn{2}{c}{\textbf{Caltech101}} & \multicolumn{2}{c}{\textbf{OxfordFlowers}} \\
    \cline{4-17}
    & & & \textbf{BF} & \textbf{AF} & \textbf{BF} & \textbf{AF} & {\makecell[c]{\textbf{Synt.} \\ \textbf{train}}} & {\makecell[c]{\textbf{Real} \\ \textbf{valid.}}} & \textbf{BF} & \textbf{AF} & \textbf{BF} & \textbf{AF} & \textbf{BF} & \textbf{AF} & \textbf{BF} & \textbf{AF}\\
    \midrule

    Discrepancy & Caltech101 & revolver & 0.96 & 0.92 & 0.856 & 0.862 & 0.016 & 0.875  & 0.558 & 0.552 & 0.517 & 0.507 &  - & -  & 0.661 & 0.632  \\
    Discrepancy & StanfordCars & 2009 Bentley Arnage Sedan & 0.692 & 0.051 & 0.557 & 0.538 & 0.922 & 0.0  & - & - & 0.517 & 0.527 &  0.857 & 0.875  & 0.661 & 0.644  \\
    \midrule
    \midrule
    No Discrepancy & Caltech101 & revolver & 0.96 & 0.16 & 0.856 & 0.855 & 0.016 & 0.0  & 0.558 & 0.524 & 0.517 & 0.48 &  - & -  & 0.661 & 0.604  \\
    No Discrepancy & StanfordCars & 2009 Bentley Arnage Sedan & 0.692 & 0.026 & 0.557 & 0.543 & 0.031 & 0.0 & - & - & 0.517 & 0.511 &  0.857 & 0.852  & 0.661 & 0.655  \\

    \bottomrule
    \end{tabular}}
    \label{table:forget_mismatch}
\end{table}

\clearpage

\twocolumn

\section{Additional Tasks}
\label{app:addtasks}

In the main paper, among the additional tasks we tested the image-image retrieval on the model after forgetting. We surprisingly found that even after forgetting the model is still able to retrieve images of the class it has forgotten starting from an input image. We speculated that in image retrieval, the model can still identify similar features and shapes of objects without actually recognizing or knowing the specific class they belong to. To confirm our hypothesis we conduct image-image retrieval on the \textbf{original} CLIP model on classes it predicts with \textbf{zero classification accuracy}. The results that confirm our hypothesis are shown in Tab. \ref{table:imageimage_additional}.

\begin{table}[!h]
 \onecolumn\caption{Image retrieval from image input results on classes with zero classification accuracy.}
    \centering
    \fontsize{7}{9}\selectfont 
    \setlength{\tabcolsep}{3pt} 
    \resizebox{0.9\textwidth}{!}{\begin{tabular}{@{}clcccc@{}} 
    \toprule 
    \textbf{Model Type} & \textbf{Class} & \textbf{Classification Accuracy} & \textbf{Precision@1} & \textbf{Precision@5} & \textbf{Precision@10} \\
    \midrule
    CLIP original & Appenzeller (StanfordDogs) & 0 & 1.0 & 0.4 & 0.2 \\
    CLIP original & Pembroke (StanfordDogs) & 0 & 1.0 & 0.6 & 0.4 \\
    CLIP original & Cardigan (StanfordDogs) & 0 & 0.0 & 0.2 & 0.2 \\
    CLIP original & 2010 Chevrolet HHR SS (StanfordCars) & 0 & 1.0 & 0.4 & 0.4 \\
    CLIP original & 2009 HUMMER H2 SUT Crew Cab (StanfordCars) & 0 & 1.0 & 0.6 & 0.7 \\
    CLIP original & english marigold (OxfordFlowers) & 0 & 1.0 & 0.8 & 0.6 \\
    CLIP original & colt's foot (OxfordFlowers) & 0 & 1.0 & 0.8 & 0.7 \\
    CLIP original & cape flower (OxfordFlowers) & 0 & 1.0 & 1.0 & 1.0 \\
    \bottomrule
    \end{tabular}}
    \label{table:imageimage_additional}
\end{table}

\onecolumn

\subsection{Full results}

\begin{table}[!h]
 \caption{Image retrieval from text input results showing precision@k for k of 1, 5 and 10 using ViT-B/16 model}
    \centering
    \fontsize{7}{9}\selectfont 
    \setlength{\tabcolsep}{3pt} 
    \resizebox{0.5\textwidth}{!}{\begin{tabular}{@{}clccc@{}} 
    \toprule 
    \textbf{Model Type} & \textbf{Class} & \textbf{Precision@1} & \textbf{Precision@5} & \textbf{Precision@10} \\
    \midrule
    CLIP original & Scotch terrier & 0.0 & 0.0 & 0.1 \\
    CLIP original & toy poodle & 1.0 & 0.8 & 0.7 \\
    CLIP original & Pekinese & 1.0 & 0.4 & 0.5 \\
    CLIP original & 2009 Spyker C8 Coupe & 1.0 & 0.8 & 0.8 \\
    CLIP original & 2010 Dodge Ram Pickup 3500 Crew Cab & 1.0 & 0.6 & 0.5 \\
    CLIP original & 2011 Ford Ranger SuperCab & 1.0 & 0.8 & 0.5 \\
    CLIP original & euphonium & 1.0 & 1.0 & 1.0 \\
    CLIP original & minaret & 1.0 & 1.0 & 1.0 \\
    CLIP original & platypus & 1.0 & 1.0 & 0.9 \\
    CLIP original & gazania & 1.0 & 1.0 & 1.0 \\
    CLIP original & tree mallow & 0.0 & 0.4 & 0.4 \\
    CLIP original & trumpet creeper & 1.0 & 0.8 & 0.6 \\
    CLIP original Mean & - & 0.833 & 0.717 & 0.667 \\
    \midrule
    CLIP forget & Scotch terrier & 0.0 & 0.4 & 0.4 \\
    CLIP forget & toy poodle & 0.0 & 0.0 & 0.1 \\
    CLIP forget & Pekinese & 0.0 & 0.0 & 0.2 \\
    CLIP forget & 2009 Spyker C8 Coupe & 1.0 & 0.8 & 0.8 \\
    CLIP forget & 2010 Dodge Ram Pickup 3500 Crew Cab & 0.0 & 0.0 & 0.1 \\
    CLIP forget & 2011 Ford Ranger SuperCab & 1.0 & 0.6 & 0.4 \\
    CLIP forget & euphonium & 1.0 & 1.0 & 0.6 \\
    CLIP forget & minaret & 1.0 & 1.0 & 0.9 \\
    CLIP forget & platypus & 1.0 & 1.0 & 0.5 \\
    CLIP forget & gazania & 1.0 & 0.2 & 0.4 \\
    CLIP forget & tree mallow & 0.0 & 0.0 & 0.2 \\
    CLIP forget & trumpet creeper & 0.0 & 0.2 & 0.2 \\
    CLIP forget Mean & - & \textbf{0.5} & \textbf{0.433} & \textbf{0.4} \\
    \bottomrule
    \end{tabular}}
    \label{table:imagetext_retr_vit}
\end{table}

\begin{table}[!h]
 \caption{Image retrieval from image input results showing precision@k for k of 1, 5 and 10 using ViT-B/16 model}
    \centering
    \fontsize{7}{9}\selectfont 
    \setlength{\tabcolsep}{3pt} 
    \resizebox{0.5\textwidth}{!}{\begin{tabular}{@{}clccc@{}} 
    \toprule 
    \textbf{Model Type} & \textbf{Class} & \textbf{Precision@1} & \textbf{Precision@5} & \textbf{Precision@10} \\
    \midrule
    CLIP original & Scotch terrier & 0.0 & 0.2 & 0.3 \\
    CLIP original & toy poodle & 1.0 & 0.4 & 0.5 \\
    CLIP original & Pekinese & 0.0 & 0.0 & 0.0 \\
    CLIP original & 2009 Spyker C8 Coupe & 1.0 & 0.4 & 0.3 \\
    CLIP original & 2010 Dodge Ram Pickup 3500 Crew Cab & 0.0 & 0.0 & 0.1 \\
    CLIP original & 2011 Ford Ranger SuperCab & 0.0 & 0.2 & 0.4 \\
    CLIP original & euphonium & 1.0 & 0.8 & 0.9 \\
    CLIP original & minaret & 1.0 & 1.0 & 1.0 \\
    CLIP original & platypus & 1.0 & 0.6 & 0.5 \\
    CLIP original & gazania & 1.0 & 1.0 & 0.7 \\
    CLIP original & tree mallow & 1.0 & 1.0 & 0.6 \\
    CLIP original & trumpet creeper & 1.0 & 1.0 & 0.7 \\
    CLIP original Mean & - & \textbf{0.667} & \textbf{0.55} & 0.5 \\
    \midrule
    CLIP forget & Scotch terrier & 0.0 & 0.2 & 0.3 \\
    CLIP forget & toy poodle & 1.0 & 0.4 & 0.5 \\
    CLIP forget & Pekinese & 0.0 & 0.0 & 0.0 \\
    CLIP forget & 2009 Spyker C8 Coupe & 1.0 & 0.4 & 0.2 \\
    CLIP forget & 2010 Dodge Ram Pickup 3500 Crew Cab & 0.0 & 0.0 & 0.1 \\
    CLIP forget & 2011 Ford Ranger SuperCab & 0.0 & 0.4 & 0.3 \\
    CLIP forget & euphonium & 1.0 & 0.8 & 0.9 \\
    CLIP forget & minaret & 1.0 & 1.0 & 0.9 \\
    CLIP forget & platypus & 1.0 & 0.6 & 0.5 \\
    CLIP forget & gazania & 1.0 & 1.0 & 0.7 \\
    CLIP forget & tree mallow & 1.0 & 1.0 & 0.7 \\
    CLIP forget & trumpet creeper & 1.0 & 1.0 & 0.7 \\
    CLIP forget Mean & - & \textbf{0.667} & 0.567 & \textbf{0.483} \\
    \bottomrule
    \end{tabular}}
    \label{table:imagetext_retr_vit}
\end{table}

\begin{table}[!h]
 \caption{Image retrieval from text input results showing precision@k for k of 1, 5 and 10 using RN50 model}
    \centering
    \fontsize{7}{9}\selectfont 
    \setlength{\tabcolsep}{3pt} 
    \resizebox{0.5\textwidth}{!}{\begin{tabular}{@{}clccc@{}} 
    \toprule 
    \textbf{Model Type} & \textbf{Class} & \textbf{Precision@1} & \textbf{Precision@5} & \textbf{Precision@10} \\
    \midrule
    CLIP original & Scotch terrier & 1.0 & 0.2 & 0.2 \\
    CLIP original & toy poodle & 1.0 & 0.6 & 0.5 \\
    CLIP original & Pekinese & 1.0 & 0.8 & 0.6 \\
    CLIP original & 2009 Spyker C8 Coupe & 1.0 & 0.6 & 0.5 \\
    CLIP original & 2010 Dodge Ram Pickup 3500 Crew Cab & 1.0 & 0.2 & 0.2 \\
    CLIP original & 2011 Ford Ranger SuperCab & 0.0 & 0.2 & 0.2 \\
    CLIP original & euphonium & 1.0 & 1.0 & 1.0 \\
    CLIP original & minaret & 1.0 & 1.0 & 1.0 \\
    CLIP original & platypus & 1.0 & 1.0 & 0.6 \\
    CLIP original & gazania & 1.0 & 1.0 & 1.0 \\
    CLIP original & tree mallow & 0.0 & 0.8 & 0.7 \\
    CLIP original & trumpet creeper & 1.0 & 0.8 & 0.5 \\
    CLIP original Mean & - & 0.833 & 0.683 & 0.583 \\
    \midrule
    CLIP forget & Scotch terrier & 0.0 & 0.0 & 0.0 \\
    CLIP forget & toy poodle & 1.0 & 0.2 & 0.1 \\
    CLIP forget & Pekinese & 0.0 & 0.0 & 0.0 \\
    CLIP forget & 2009 Spyker C8 Coupe & 0.0 & 0.8 & 0.5 \\
    CLIP forget & 2010 Dodge Ram Pickup 3500 Crew Cab & 0.0 & 0.2 & 0.3 
    \\
    CLIP forget & 2011 Ford Ranger SuperCab & 0.0 & 0.0 & 0.0 \\
    CLIP forget & euphonium & 0.0 & 0.8 & 0.8 \\
    CLIP forget & minaret & 0.0 & 0.4 & 0.2 \\
    CLIP forget & platypus & 0.0 & 0.2 & 0.2 \\
    CLIP forget & gazania & 0.0 & 0.0 & 0.0 \\
    CLIP forget & tree mallow & 0.0 & 0.2 & 0.2 \\
    CLIP forget & trumpet creeper & 0.0 & 0.0 & 0.0 \\
    CLIP forget Mean & - & \textbf{0.08} & \textbf{0.23} & \textbf{0.191} \\
    \bottomrule
    \end{tabular}}
    \label{table:imagetext_retr_vit}
\end{table}

\begin{table}[!h]
 \caption{Image retrieval from image input results showing precision@k for k of 1, 5 and 10 using RN50 model}
    \centering
    \fontsize{7}{9}\selectfont 
    \setlength{\tabcolsep}{3pt} 
    \resizebox{0.5\textwidth}{!}{\begin{tabular}{@{}clccc@{}} 
    \toprule 
    \textbf{Model Type} & \textbf{Class} & \textbf{Precision@1} & \textbf{Precision@5} & \textbf{Precision@10} \\
    \midrule
    CLIP original & Scotch terrier & 0.0 & 0.0 & 0.0 \\
    CLIP original & toy poodle & 0.0 & 0.0 & 0.1 \\
    CLIP original & Pekinese & 0.0 & 0.0 & 0.0 \\
    CLIP original & 2009 Spyker C8 Coupe & 1.0 & 0.4 & 0.3 \\
    CLIP original & 2010 Dodge Ram Pickup 3500 Crew Cab & 0.0 & 0.0 & 0.0 \\
    CLIP original & 2011 Ford Ranger SuperCab & 0.0 & 0.2 & 0.3 \\
    CLIP original & euphonium & 0.0 & 0.4 & 0.2 \\
    CLIP original & minaret & 1.0 & 0.8 & 0.6 \\
    CLIP original & platypus & 0.0 & 0.2 & 0.2 \\
    CLIP original & gazania & 1.0 & 0.8 & 0.6 \\
    CLIP original & tree mallow & 1.0 & 0.8 & 0.8 \\
    CLIP original & trumpet creeper & 1.0 & 0.8 & 0.7 \\
    CLIP original Mean & - & 0.417 & \textbf{0.367} & \textbf{0.317} \\
    \midrule
    CLIP forget & Scotch terrier & 0.0 & 0.0 & 0.0 \\
    CLIP forget & toy poodle & 0.0 & 0.0 & 0.1 \\
    CLIP forget & Pekinese & 0.0 & 0.0 & 0.0 \\
    CLIP forget & 2009 Spyker C8 Coupe & 1.0 & 0.4 & 0.3 \\
    CLIP forget & 2010 Dodge Ram Pickup 3500 Crew Cab & 0.0 & 0.0 & 0.1 \\
    CLIP forget & 2011 Ford Ranger SuperCab & 0.0 & 0.2 & 0.3 \\
    CLIP forget & euphonium & 0.0 & 0.4 & 0.2 \\
    CLIP forget & minaret & 0.0 & 0.8 & 0.6 \\
    CLIP forget & platypus & 0.0 & 0.2 & 0.2 \\
    CLIP forget & gazania & 1.0 & 0.8 & 0.6 \\
    CLIP forget & tree mallow & 1.0 & 0.8 & 0.8 \\
    CLIP forget & trumpet creeper & 1.0 & 0.8 & 0.7 \\
    CLIP forget Mean & - & \textbf{0.333} & \textbf{0.367} & 0.325 \\
    \bottomrule
    \end{tabular}}
    \label{table:imagetext_retr_vit}
\end{table}

\clearpage
\twocolumn

\section{Additional Figures and Implementation Details}
\label{app:add_impl}

\paragraph{Implementation Details}
For CLIP with ResNet50 visual encoder, we observed that allowing all parameters in the vision encoder to be chosen for update during the forgetting process results in poor performance on other classes. Therefore, we restrict parameter updates to the attention layers in the RN50 vision encoder while all parameters in the text encoder are eligible for the update selection. For the ViT model all the parameters are allowed to vary. Fig. \ref{fig:layers_update} illustrates the top 25 most frequently updated layers of the RN50 model. We generate 64 synthetic samples and stop the forgetting process when their accuracy goes below 0.1. \\
We use Adam optimizer with learning rate of 5e-05, weight decay of 0.2 and betas parameters of 0.9 and 0.98. We set the number of perturbed samples $N$ to 25. Initial $\sigma$ is 0.1 increasing to 2. Initially, layers we allow to vary are 5 for both visual and textual encoders and increasing to 8 and 20 respectively if we need more forgetting according to the remaining synthetic samples accuracy. 
All the experiments are run on a single NVIDIA GeForce RTX 3090 with 24GB of memory.

\paragraph{Additional Figures}
In the vision encoder, the weights of the values, queries, and output projections are updated most frequently, whereas in the text encoder the MLP output projection weights are updated most often.

\begin{figure}[!h]
\onecolumn \centering
\footnotesize
\includegraphics[width=0.60\textwidth]{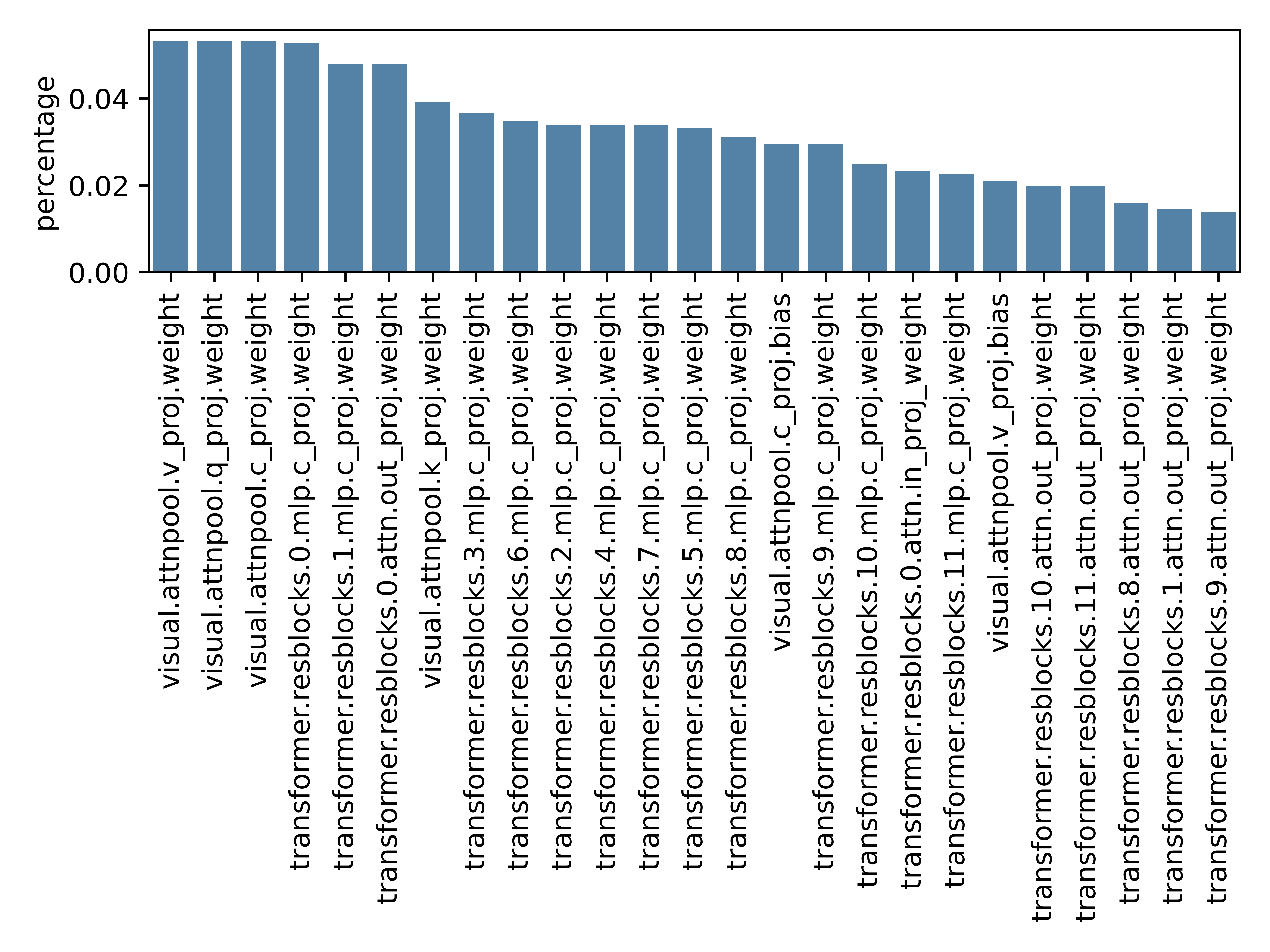}
\caption{\textbf{Selected layers for forgetting.} The figure shows the top 25 most frequent updated layers during forgetting process across selected classes and datasets. }
\label{fig:layers_update}
\end{figure}

}{%
}

\end{document}